\documentclass[11pt]{article}

\usepackage[final]{acl}

\usepackage{times}
\usepackage{latexsym}
\usepackage{makecell}
\usepackage{algorithm}
\usepackage{algpseudocode}
\usepackage{amsmath}
\usepackage{tcolorbox}
\usepackage{fancyvrb}
\usepackage{listings}
\usepackage[table]{xcolor}
\usepackage{xcolor}
\usepackage{subcaption}
\usepackage{multirow}
\usepackage{multicol}
\usepackage{booktabs}
\usepackage{amssymb}
\usepackage{enumitem}
\usepackage[T1]{fontenc}

\usepackage[utf8]{inputenc}

\usepackage{microtype}

\usepackage{inconsolata}

\usepackage{graphicx}

\newcommand{\upA}[1]{\textcolor{blue!80!black}{\tiny$\uparrow$#1}}
\newcommand{\downA}[1]{\textcolor{red}{\tiny$\downarrow$#1}}
        
\newcommand{\upB}[1]{\textcolor{teal}{\tiny\boldmath$\Uparrow$#1}}

\newcommand{\gap}{\hspace{1pt}}
 
\title{VinciCoder: Unifying Multimodal Code Generation via Coarse-to-fine Visual Reinforcement Learning}

\author{Xuanle Zhao$^\text{1,2 }$\thanks{Equal contribution.}, Deyang Jiang$^\text{1 }$\footnotemark[1], Zhixiong Zeng$^\text{1 }$\thanks{Corresponding author.}, Lei Chen$^\text{1}$, Haoyue Yang$^\text{2}$, Haibo Qiu$^\text{1}$,\\ 
\textbf{Jing Huang$^\text{1}$, Yufeng Zhong$^\text{1}$, Liming Zheng$^\text{1}$, Yilin Cao$^\text{1}$, Lin Ma$^\text{1 }$\footnotemark[2]} \\
\textbf{\textsuperscript{1}} Meituan
\textbf{\textsuperscript{2}} Institute of Automation, Chinese Academy of Sciences
\\
  \texttt{zengzhixiong@meituan.com, forest.linma@gmail.com} \\
}

\begin{document}
\maketitle
\begin{abstract}
While recent specialized multimodal code generation models excel in tasks like chart-to-code generation, their reliance on single-task training limits generalization and hinders the development of \textbf{VI}sio\textbf{N} \textbf{C}ode \textbf{I}ntelligence. In this work, we introduce \textbf{VinciCoder}, a unified framework designed for generalized multimodal code generation.
We first curate a large-scale SFT corpus comprising 1.3M direct generation pairs and 300k visual-based refinement tasks. This composition fosters self-refinement capabilities, enabling the model to directly rectify code to align with input images.
Subsequently, we propose coarse-to-fine Visual Reinforcement Learning (ViRL) to overcome the brittleness of textual metrics in handling semantically equivalent but syntactically diverse code. By quantifying visual similarity across multi-scale patches, ViRL provides an implementation-agnostic reward mechanism that ensures high-fidelity alignment between rendered outputs and input visuals.
Extensive experimental results across diverse benchmarks demonstrate that VinciCoder achieves superior performance, while comprehensive ablation studies validate the effectiveness of our proposed ViRL strategy. The data, code and model are available at \url{https://github.com/DocTron-hub/VinciCoder}.
\end{abstract}

\section{Introduction}
\label{sec:intro}
Recent advancements in Large Language Models (LLMs), such as Gemini-2.5 \cite{comanici2025gemini} and Qwen3-Coder \cite{yang2025qwen3}, have achieved significant breakthroughs in code generation, demonstrating a robust capability to follow complex instructions and synthesize executable code across various programming languages.
Beyond textual descriptions, a growing body of research \cite{wan2024automatically, jiang2025screencoder} explores multimodal code generation from visual inputs like charts and webpages, which are inherently more information-dense than natural language and thus present greater challenges for model comprehension and synthesis.

\begin{figure}[t]
    \centering
    \includegraphics[width=0.48\textwidth]{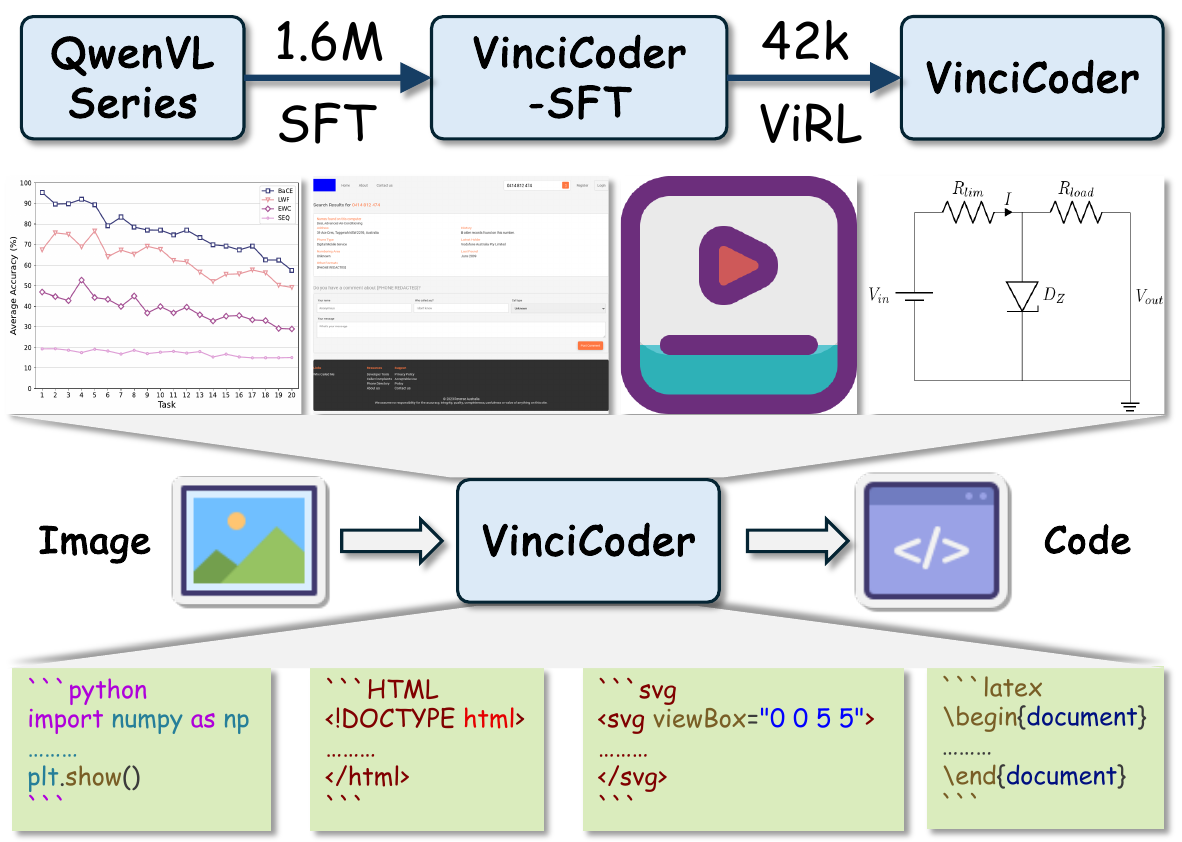}
    \vspace{-20pt}
    \caption{The VinciCoder Framework. Built upon the Qwen-VL series via a progressive SFT-ViRL pipeline, VinciCoder bridges visual perception and code synthesis across diverse modalities.}
    \label{fig:intro}
    \vspace{-15pt}
\end{figure}

However, current research predominantly develops task-specific vision-language models (VLMs) such as ChartCoder \cite{zhao2025chartcoder} for charts and Web2Code \cite{yun2024web2code} for web-to-HTML. While effective in their respective domains, these models rely on narrow training scopes that restrict generalization and purely supervised fine-tuning (SFT) that fails to ensure both high code executability and visual fidelity. Furthermore, while Reinforcement Learning with Verifiable Reward (RLVR) has seen broad cross-domain adoption, its application to code generation is hindered by brittle, rule-based rewards that induce instability in diverse, open-ended scenarios due to their sensitivity to semantically equivalent but syntactically diverse code. Consequently, developing a unified multimodal code generation framework \cite{jiang2025viscodex, sun2025januscoder} integrated with style-agnostic RL has emerged as a vital yet challenging research direction.

To address these challenges, we introduce VinciCoder, a unified VLM that employs a two-stage training strategy for multimodal code generation. In the SFT stage, we curate a large-scale corpus of 1.3M direct generation samples and pioneer the integration of a non-trivial visual-based refinement task using 300k additional samples.
Unlike conventional editing, this task requires the model to rectify flawed code snippets containing logical errors or partial renderings, ensuring their visual outputs precisely align with target images. By incorporating this data, we equip VinciCoder with multi-turn debugging and refining capabilities to iteratively improve code quality.
Subsequently, we propose Visual Reinforcement Learning (ViRL) to enhance code executability and visual fidelity.
By pivoting the reward mechanism to the visual domain, ViRL utilizes direct perceptual similarity as a style-agnostic signal, ensuring stable optimization that is indifferent to specific coding implementations.
To robustly compare multi-resolution images, our ViRL framework employs a coarse-to-fine reward mechanism that simultaneously captures global structural layouts from downsampled views and fine-grained local details from segmented patches.

We conduct extensive experiments across diverse multimodal benchmarks, comparing VinciCoder against leading contemporary models. The results demonstrate the effectiveness of our two-stage strategy, as VinciCoder-SFT already establishes a strong baseline that surpasses existing models. The subsequent ViRL further enhances performance, setting a new frontier across all tasks. Furthermore, ablation studies on cold-start ViRL settings (w/o SFT) confirm that our proposed ViRL strategy consistently improves model performance independently. Beyond direct generation, the visual refinement task empowers VinciCoder to debug and refine code snippets using ground-truth visual feedback. To our knowledge, VinciCoder is the first unified model to leverage RL for domain-agnostic visual fidelity in multimodal code generation.
\begin{itemize}[leftmargin=*] 
\item We introduce VinciCoder, a unified VLM for generalized multimodal code generation. By pioneering a non-trivial visual-based refinement task, we empower the model with self-refinement capabilities to rectify code discrepancies.
\item We propose coarse-to-fine ViRL, a framework that enables stable optimization for semantically equivalent but syntactically diverse code by shifting from brittle textual rewards to implementation-agnostic visual similarity.
\item Extensive evaluation results across diverse benchmarks demonstrate the superior performance of VinciCoder. Ablation results further demonstrate that our method significantly enhances both code executability and visual fidelity.
\end{itemize}

\section{Related Works}
\subsection{MLLMs for Code Generation}
Multimodal code generation has gained significant traction, particularly in synthesizing code for visual artifacts such as charts, webpages, SVGs, and scientific plots.
Current research in multimodal code generation spans several specialized domains. In charts and scientific plots, works like ChartCoder \cite{zhao2025chartcoder} and MathCoder-VL \cite{wang2025mathcoder} focus on large-scale corpora and real-world datasets (e.g., Datikz \cite{belouadi2025tikzero}), while others integrate RL \cite{chen2025breaking} to enhance visual fidelity. Similarly, web-to-code research leverages both synthetic pairs \cite{yun2024web2code} and real-world data \cite{gui2025webcode2m}, with models like LayoutCoder \cite{wu2025mllm} introducing layout-aware frameworks for structural accuracy. In the SVG domain, StarVector \cite{rodriguez2025starvector} and OmniSVG \cite{yang2025omnisvg} utilize paired icons and illustrations to bridge the gap between images and scalable graphics. Furthermore, approaches like Cosyn \cite{yang2025scaling} utilize LLMs to synthesize code across multiple visual domains, enabling the rendering of informative images from textual descriptions.
However, these methods are task-specific, as they are constrained to homogeneous visual patterns and singular programming languages. While recent research has shifted toward unified models \cite{jiang2025viscodex, sun2025januscoder}, these efforts rely almost exclusively on SFT, which is insufficient for ensuring both code executability and visual fidelity.

\begin{figure*}[t]
    \centering
    \includegraphics[width=0.98\textwidth]{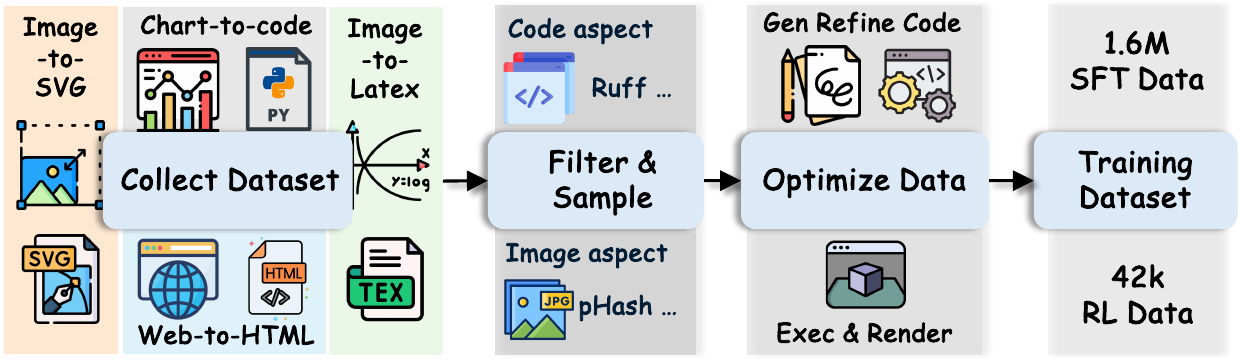}
    \vspace{-5pt}
    \caption{Overview of the VinciCoder data construction pipeline. Following initial filtering and diversity-aware sampling of open-source datasets, the corpus is enhanced through two parallel streams: refining existing samples via execution and optimization, and generating novel samples for the visual refinement task. This dual strategy produces high-quality data pairs for both SFT and RL training stages.}
    \label{fig:result}
    \vspace{-10pt}
\end{figure*}

\subsection{Reinforcement Learning for MLLM}
Inspired by the success of Group Relative Policy Optimization (GRPO) in post-training LLMs \cite{shao2024deepseekmath, guo2025deepseek}, RL has garnered significant attention from the research community.
Recently, a growing body of work has focused on applying RL to MLLMs to enhance their vision-language reasoning capabilities. Pioneering efforts, such as Vision-R1 \cite{huang2025vision}, VLM-R1 \cite{shen2025vlm}, and R1-OneVision \cite{yang2025r1}, initially utilize Chain-of-Thought (CoT) data for SFT to establish a reasoning baseline and employ RL to refine the model policy towards generating coherent answers.
This RL framework has proven highly versatile, with successful adaptations to a range of fundamental visual tasks, including grounding \cite{zhou2025gui, liu2025visual}, question answering (QA) \cite{tan2025reason, chen2025chart}, and segmentation \cite{liu2025seg}. Besides constructing rewards from generated text contents, another research direction involves leveraging visual feedback to formulate reward functions. For instance, RRVF \cite{chen2025learning} and RLRF \cite{rodriguez2025rendering} utilize MLLMs and pretrained Vision Transformers (ViTs) to score generated images, thereby providing a reward signal for RL training.

\section{Method}
\subsection{Task Definition}
The standard approach for multimodal code generation is to generate code from visual and textual inputs directly. Given an $\operatorname{Image}$ and a $\operatorname{Text}$ instruction, an MLLM is tasked with generating the corresponding $\operatorname{Code}$.
\begin{equation}
\small
\operatorname{Code} = \operatorname{MLLM}(\operatorname{Image}, \operatorname{Text})
\end{equation}
Besides direct generation, we introduce a visual-based refinements task, which tasks the model with refining an initial, potentially flawed code draft $\operatorname{Code}_{\operatorname{draft}}$ to the refined version $\operatorname{Code}_{\operatorname{refined}}$. 
\begin{equation}
\small
\operatorname{Code}_{\operatorname{refined}} = \operatorname{MLLM}(\operatorname{Image}, \operatorname{Code}_{\operatorname{draft}}, \operatorname{Text})
\end{equation}

\subsection{Data Construction}
To construct large-scale training data, we curate datasets from various open-source datasets and generate missing image-code types. All code is re-executed to render the corresponding images. 

\subsubsection{SFT Data}
\textbf{Chart-to-code.} 
For our chart-to-code task, we curate training data from ChartCoder \cite{zhao2025chartcoder} and MSRL \cite{chen2025breaking}, starting with rectifying syntax errors in Chart2Code-160k via automated linters.To ensure diversity, we extract 100k distinct samples from MSRL using perceptual hashing (pHash) and mini-batch K-means clustering. We further construct the code refinement dataset by employing a specialized VLM, trained on our initial corpus, to generate first-pass code for 100k non-overlapping MSRL samples. This dataset pairs imperfect generations with their corresponding target images and ground-truth code, facilitating the training of multi-turn debugging capabilities. We provide the refinement data format in Figure~\ref{fig:refine_task}.

\textbf{Web-to-HTML.} For the web-to-HTML task, we curate training data from MCD \cite{jiang2025viscodex}, Web2M \cite{gui2025webcode2m}, and Web2Code \cite{yun2024web2code}. For the Web2M collection, we filter for English-language entries and exclude samples containing hyperlinks or embedded images, yielding a refined dataset of 60k entries. Following a pipeline consistent with our chart-to-code approach, we train a dedicated web-to-code VLM on the combined MCD and filtered Web2M data. This model is then used to generate initial code for 100k instances sampled from the Web2Code dataset to form the final refinement set.

\textbf{Image-to-SVG.} For the image-to-SVG task, our primary corpus comprises 360k image-SVG pairs from the UniSVG ISVGEN subset \cite{li2025unisvg}. To construct the refinement dataset, we employ a dedicated VLM trained on ISVGEN to generate code drafts for 100k samples from the same subset. This approach is motivated by observed training instabilities, where high and fluctuating final training loss suggests the model has not yet mastered the training set. Consequently, these samples remain non-trivial targets for refinement, allowing the model to learn error correction even on previously seen data.

\textbf{Image-to-Latex.} For the image-to-LaTeX task, we curate data from the DaTikZ-v3 \cite{belouadi2024detikzify} and Cosyn-400k \cite{yang2025scaling}. To ensure consistent and valid outputs, we standardize the code by encapsulating all instances within a standalone TikZ environment. This procedure is specifically designed to generate tightly-cropped figures, thereby preventing the model from producing full-page A4 PDF layouts. Following this, we execute each sample to validate its integrity, filtering out instances that either result in rendering errors or produce multi-page outputs.

\begin{table}[t]
\small
\setlength{\tabcolsep}{6pt}
\centering
\resizebox{0.48\textwidth}{!}{
\begin{tabular}{lcc}
\toprule
\textbf{Scientific Plots} & \textbf{Code Types} & \textbf{Statistics} \\
\midrule
Document & Latex/HTML & 71k \\
Molecule & RDKiT/Indigo & 50k \\
Diagram & Latex/HTML/Mermaid & 48k\\
Table & Latex/HTML & 32k\\
Graphic & SVG/Asymptote & 27k \\
Circuit & Latex & 10k \\
\bottomrule
\end{tabular}}
\vspace{-5pt}
\caption{Details about scientific plots and corresponding code types in the SFT dataset.}
\label{tab:scientific_plots}
\vspace{-10pt}
\end{table}

\textbf{Scientific Plots-to-code.} In addition to the aforementioned domains, we extend our investigation to complex scientific visualizations. This expansion covers a broad spectrum of graphical representations, including molecular structures, electronic schematics, general diagrams, document layouts, and tabular data. The underlying code for these figures utilizes both general-purpose languages and specialized domain-specific languages (DSLs) such as Mermaid and Asymptote. Our dataset is primarily constructed from the Cosyn-400k collection and various open-source text-to-Mermaid datasets. We further augment this corpus with 40k molecular image-code pairs, which were generated by rendering SMILES strings sourced from the USPTO database.
Table~\ref{tab:scientific_plots} provides a detailed breakdown of the dataset, summarizing the distribution of categories and their corresponding statistics.

\subsubsection{RL Data}
For the RL phase, we construct a new dataset spanning five distinct domains, ensuring complete mutual exclusivity with our SFT corpus. The curation process for each domain is detailed as follows: (i) Chart-to-code: We directly utilize the 11k image-code pairs from the second RL-stage subset of the MSRL dataset \cite{chen2025breaking}. (ii) Web-to-HTML: We sample 9k examples from Web2Code \cite{yun2024web2code} and employ Gemini-2.5-Flash \cite{comanici2025gemini} to refine the HTML code, using varied reference tags and diverse instructions to enhance the visual complexity and quality of the generated webpages. (iii) Image-to-SVG: Using our trained SFT model as a quality filter, we curate the SVG-icons dataset \cite{rodriguez2025starvector} by identifying hard examples. We retain 10k high-fidelity samples that the current model fails to reconstruct accurately, specifically targeting complex cases that remain challenging for generative synthesis.
(iv) Image-to-LaTeX: We extract samples from the arxiv-woc-680k category of the ImgCode-8.6M dataset \cite{wang2025mathcoder}. To rectify issues such as inconsistent output scaling and low content diversity, we implement a hybrid filtering pipeline that combines rule-based heuristics with model-based verification, yielding a final curated set of 10k examples. (v) Chemical Images: We synthesize a dedicated collection of 2k molecular images by rendering unique SMILES strings through both RDKit \cite{landrum2013rdkit} and Indigo \cite{IndigoToolkit}, ensuring diverse rendering styles for the RL phase.

\begin{table}[t]
\setlength{\tabcolsep}{8pt}
\small
\centering
\resizebox{0.48\textwidth}{!}{
\begin{tabular}{lccc}
\toprule
\multirow{2}{*}{\textbf{Image}} & \multirow{2}{*}{\textbf{Code}} & 
\multicolumn{2}{c}{\textbf{Data Statistics}} \\
\cmidrule{3-4}
& &\textbf{SFT Data} & \textbf{RL Data} \\
\midrule
Chart &Python & 412k & 11k \\
Webpage & HTML & 355k & 9k \\
Image & SVG & 463k & 10k \\
Image & Latex & 108k & 10k \\
Scientific Plots & Multiple & 238k & 2k \\
\bottomrule
\end{tabular}}
\vspace{-5pt}
\caption{Details about SFT and RL data statistics.}
\label{tab:datasets}
\vspace{-10pt}
\end{table}

\begin{figure*}[t]
    \centering
    \includegraphics[width=0.98\textwidth]{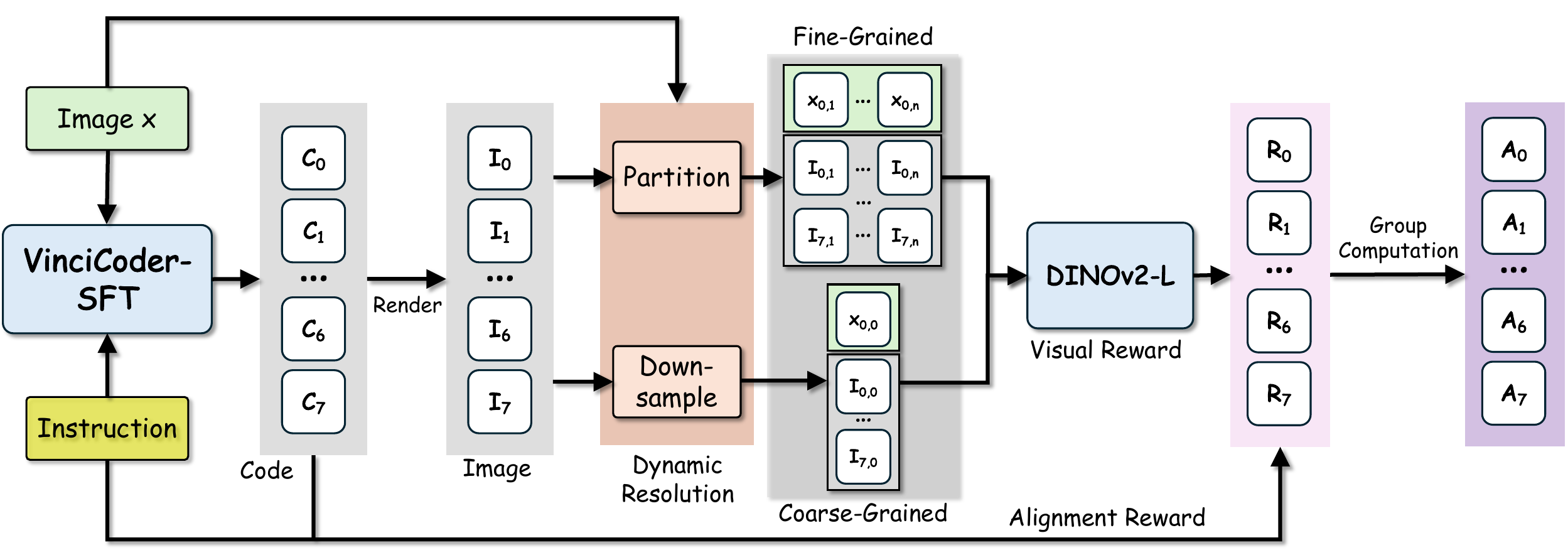}
    \vspace{-5pt}
    \caption{Overview of the ViRL strategy. During training, the alignment reward assesses the structural and linguistic consistency between the generated code and input instructions. Subsequently, for executable rollouts, the coarse-to-fine visual reward is computed by comparing DINOv2 embeddings of global thumbnails and local patches against the target image to evaluate visual alignment.}
    \label{fig:rl_algo}
    \vspace{-10pt}
\end{figure*}
The comprehensive composition of our training datasets is detailed in Table~\ref{tab:datasets}. Specifically, the chart-to-code, web-to-HTML, and image-to-SVG subsets are augmented with 100k, 92k, and 103k refinement samples, respectively, while the rest consists exclusively of direct generation data.

\subsection{Model Training}
\subsubsection{Supervised Finetuning (SFT)}
The SFT phase is essential for establishing syntactic priors, preventing the model from over-optimizing for visual fidelity at the expense of functional correctness. In niche or less-trained tasks, such as chemical tasks, direct ViRL without this initialization often triggers reward hacking, prioritizing perception similarity over code integrity. Consequently, we first employ a standard autoregressive objective to train the model
\begin{equation}
\small
\mathcal{L}(\theta):=-\mathbb{E}_{(x, y) \sim \mathcal{D}_{\text {SFT}}} \sum_{t=1}^T \log P\left(y_t \mid x, y_{<t} ; \theta\right),
\end{equation}
\noindent$(x, y)$ is the query and target response.

\subsubsection{Visual Reinforcement Learning (ViRL)}
Although SFT achieves strong performance on text-based similarity, this approach faces fundamental limitations in capturing visual-semantic alignment. First, the next-token prediction objective is inherently local and provides no supervisory signals for global functional properties such as code executability. Second, SFT lacks explicit visual grounding as the model remains blind to the rendered output during training. This deficiency is critical because the code-to-visual mapping is highly sensitive, where minor syntax perturbations can induce catastrophic shifts in the rendered images. 

To address these issues, we propose ViRL, a reinforcement learning strategy that incorporates coarse-to-fine visual rewards and alignment rewards to optimize visual fidelity and instruction-following consistency, respectively. During the ViRL stage, we employ the Group Relative Policy Optimization (GRPO) \cite{shao2024deepseekmath} to optimize the model. The total reward $R$ is a weighted combination of the coarse-to-fine visual reward $R_v$ and the language alignment reward $R_l$:
\begin{equation}
\small
R = \omega_v R_v + \omega_l R_l
\end{equation}
where $\omega_v$ and $\omega_l$ are hyperparameters controlling the relative importance. Figure~\ref{fig:rl_algo} illustrates the integrated framework of our ViRL strategy.


\begin{table*}[t]
\setlength{\tabcolsep}{2pt}
\small
\centering
\resizebox{\textwidth}{!}{
\begin{tabular}{l|ccc|ccc|cc|cc|cc}
\toprule
\multirow{2}{*}{Model} & \multicolumn{3}{c|}{\textbf{ChartMimic-direct-v2}} & \multicolumn{3}{c|}{\textbf{UniSVG-ISVGEN}} & \multicolumn{2}{c|}{\textbf{Design2Code}} & 
\multicolumn{2}{c|}{\textbf{Image2Struct-plot}} & \multicolumn{2}{c}{\textbf{ChemDraw}} \\
\cmidrule{2-13}
   & Exec.Rate & Low-L  & High-L & Low-L  & High-L  & Score & Low-L & High-L & Ren.Succ. & EMS & Exec.Rate & Tani.Sim.\\ 
\midrule

\multicolumn{13}{l}{\textbf{Closed-Source Models}} \\ 
\midrule
\rowcolor[HTML]{DCDCDC} Gemini-2.5-Pro & 97.3 & 88.7 & 83.8 & 53.6 & 80.3 & 69.6  & 90.8 & 91.4 & 74.3 & 52.5 &77.3& 2.8\\
\rowcolor[HTML]{DCDCDC} Claude-4.5-Sonnet & 97.8 & 89.6 & 82.9& 61.0 & 83.4 & 74.6 & 90.4 & 90.8& 72.7 & 50.2 & 95.3 & 41.7\\
\rowcolor[HTML]{DCDCDC} GPT-5 & 94.8 & 81.9 &  78.3 & 60.8 & 88.3 & 77.3 & 90.6 & 91.0 & 78.7 & 57.4 & 93.8 & 52.1\\
\midrule
\multicolumn{13}{l}{\textbf{Open-Source Models}} \\
\midrule
Qwen2.5-VL-72B & 88.5 & 72.7 & 79.1 & 47.7 & 76.0 & 64.7 & 86.9 & 88.7 & 62.0 & 41.7 & 75.8 & 28.0 \\
InternVL3.5-38B & 79.0 &60.0 & 71.8 & 51.9 & 77.3 & 67.1& 87.8 & 88.4 & 72.6 & 49.5 &55.5 & 31.4 \\
Qwen3-VL-32B  & 83.0 & 66.9 & 77.5 & 68.0 & 86.0 & 78.8 & \textbf{88.6} & \textbf{89.8} & 75.7 & 53.3 & 37.5 & 48.8 \\
InternVL3.5-14B & 73.2 & 52.8 & 55.4 & 52.0 & 75.0 & 65.9 & 86.1 & 87.8 & 73.0 & 50.2  & 71.9 & 39.3 \\
InternVL3-14B  & 72.3 & 51.3 & 54.1 & 51.4 & 75.5 & 65.8 & 85.8 & 87.5 & 73.3 & 52.2 & 71.1  & 40.2 \\
InternVL3.5-8B & 66.7 & 46.0 & 48.3 & 55.0 & 78.0 & 68.6 & 85.8 & 87.3 & 58.3 & 40.5 &49.2 & 7.8 \\
InternVL3-8B & 63.3 & 43.8 & 46.1 & 54.5 & 77.4 & 68.2 & 85.3 & 87.6 & 57.7& 38.6 & 42.2 & 6.2 \\
JanusCoderV-7B & 80.6 & 65.8 & 72.7 &  63.6 & 70.1 & 67.5 & 85.7 & 87.8 & - & - & - & - \\
\midrule
Qwen2.5-VL-7B  & 68.7 & 42.2 & 40.1 &  47.5 & 73.8 & 63.3 & 83.4 & 87.6 & 42.7 & 25.5 &21.1 & 11.7 \\
\rowcolor[HTML]{D7F6FF} VinciCoder-7B  & 91.2 & 78.3 & 79.8 & 77.0 & 92.0 & 86.0 & 88.2 & 89.1  & \textbf{84.7} & \textbf{60.9} & 87.5 & 56.0 \\
\midrule
Qwen3-VL-8B & 78.3 & 62.5 & 67.8 & 53.0 & 77.0 & 67.4 & 85.5  & 87.2 & 47.7 & 33.0 & 78.9 & 41.2 \\
\rowcolor[HTML]{D7F6FF} VinciCoder-8B & \textbf{91.6} & \textbf{78.9} & \textbf{80.6} & \textbf{77.1} & \textbf{94.1} & \textbf{87.3} & 88.4 & 89.3 & 77.3 & 57.8 & \textbf{88.3} & \textbf{62.6}\\
\bottomrule 
\end{tabular}
}
\vspace{-5pt}
\caption{Main results on multimodal code generation benchmarks. Gray and blue rows represent closed-source baselines and VinciCoder, respectively. The best performance among open-source models is in \textbf{bold}. JanusCoderV-7B results for Image2Struct and ChemDraw are omitted due to a lack of task-specific training.}
\label{tab:main_results}
\vspace{-10pt}
\end{table*}

\textbf{Coarse-to-fine Visual Reward $R_v$.}
Directly downsampling high-resolution images is a suboptimal approach for computing visual similarity, as it inherently discards fine-grained details and may mask significant visual discrepancies. To mitigate this, we propose a coarse-to-fine visual reward mechanism $R_v$. Given a rendered image $I_r$ and its corresponding ground-truth $I_s$, we first align their dimensions by resizing the former to the latter. The mechanism then evaluates similarity across two complementary scales:
(1) Fine-grained scale, which employs a dynamic partitioning strategy to divide the image into non-overlapping $448 \times 448$ patches while preserving the original aspect ratio; and (2) Coarse-grained scale, where a downsampled thumbnail of the entire image is generated to provide global-level context.
Similarity scores are computed between corresponding tiles and thumbnails using DINOv2 \cite{oquab2023dinov2}. The final visual reward is the arithmetic mean of the global score and all patch-based scores:
\begin{equation}
\small
R_v = \frac{1}{N+1} \left( R_{\operatorname{global}} + \sum_{i=1}^N R_{\operatorname{fine},i} \right)
\end{equation}
where $N$ denotes the number of fine-grained patches. In instances where the generated code fails to render successfully, $R_v$ is assigned a value of 0. This robust design ensures that fine and coarse rewards naturally converge for low-resolution images, enabling a unified pipeline to handle diverse resolutions.

\textbf{Alignment Reward $R_l$.}
To address the issue where the SFT model generates code in a language inconsistent with the prompt, we introduce an alignment reward to improve instruction fidelity. This reward is determined by checking the consistency between the target language specified in the instruction and the language identifier extracted from the generated code blocks. We use a predefined mapping to handle aliases, such as mapping \texttt{tikz} to \texttt{latex}. A binary reward of 1 is assigned if the generated language matches a valid alias, ensuring the model follows the specified programming language.

\section{Experiments}
\subsection{Implementation Details}

We perform SFT on the curated dataset for one epoch, employing Qwen2.5-VL-7B-Instruct \cite{bai2025qwen2} and Qwen3-VL-8B-Instruct \cite{bai2025qwen3vltechnicalreport} as the base models. This stage is conducted on 24 H800 GPUs with a global batch size of 96. During the RL phase, we utilize GRPO to fine-tune the SFT model with a global batch size of 256. The hyperparameters $\omega_v$ and $\omega_l$ are set to 0.9 and 0.1, respectively. For each query, GPRO generate 8 rollouts. For resource allocation, 16 GPUs are dedicated to the policy model while 4 GPUs are reserved for reward scoring, which utilizes DINOv2-L to generate visual embeddings. 

\subsection{Evaluation Settings}
Evaluation Benchmarks. We evaluate VinciCoder against leading closed-source \cite{comanici2025gemini, anthropic_claude_4_5_sonnet, openai_gpt_5} and open-source models \cite{bai2025qwen2, zhu2025internvl3, wang2025internvl3,sun2025januscoder} across five domains. Specifically, we utilize ChartMimic \cite{yang2024chartmimic} for charts, Design2Code \cite{si2024design2code} for webpages, UniSVG \cite{li2025unisvg} for SVG, and Image2Struct \cite{roberts2024image2struct} for LaTeX generation. For molecules, we evaluate ChemDraw performance on the Cosyn-400k \cite{yang2025scaling} test set, reporting the execution rate and average Tanimoto similarity (Tani. Sim.) of the generated SMILES.


\begin{figure*}[t]
    \centering
    \includegraphics[width=0.98\textwidth]{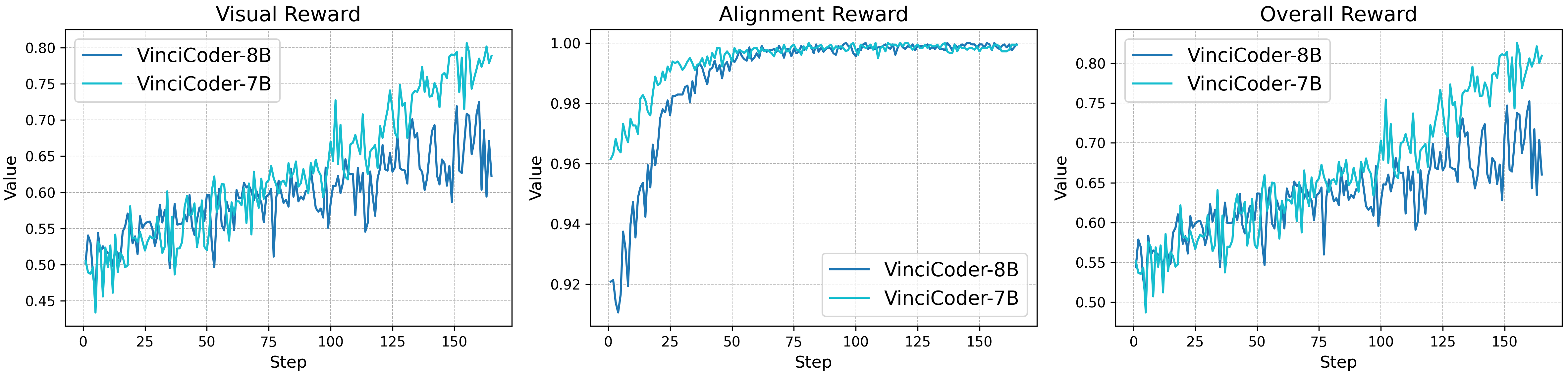}
    \vspace{-5pt}
    \caption{Reward curves during ViRL training. As training progresses, the visual reward shows a steady upward trend, while the alignment reward rapidly converges to its maximum value.}
    \label{fig:rl_result}
    \vspace{-10pt}
\end{figure*}

\subsection{Main Results}
As detailed in Table~\ref{tab:main_results}, VinciCoder establishes a new state-of-the-art among open-source solutions, outperforming existing models across the majority of benchmarks. Notably, it maintains a substantial lead over both comparable-scale competitors and the recent unified multimodal model JanusCoderV.

The performance gap between the SFT and final models validates the effectiveness of our two-stage training process. The initial SFT stage alone yields substantial gains, with VinciCoder-SFT significantly surpassing its base model across nearly all benchmarks. This result underscores the high quality of our large-scale SFT dataset. The subsequent ViRL stage further enhances performance across almost all metrics, especially in visual similarity and execution rate. This improvement stems from two mechanisms: RL penalizes invalid outputs with zero rewards to boost execution rates, while the ViRL strategy optimizes perceptual alignment between rendered and target images to improve visual similarity. We visualize reward progressions during the ViRL stage in Figure~\ref{fig:rl_result}.
In addition to generalist models, comparisons against task-specific baselines are provided in the Appendix~\ref{sec:task_specific}.

\begin{table}[t]
\centering
\small
\setlength{\tabcolsep}{2pt} 
\begin{tabular}{l|cc}
\toprule
\textbf{Dataset} & \textbf{Debug (Non-Exec.)} & \textbf{Refine (Exec.)} \\
 & Fixed / Total (Fix Rate) & 1st $\to$ 2nd ($\Delta$ High-L) \\
\midrule
ChartMimic & 12 / 66 (18.2\%) & 88.6 $\to$ 89.8 (+1.2) \\
UniSVG     & 8 / 85 (9.4\%)  & 91.0 $\to$ 92.3 (+1.3) \\
\bottomrule
\end{tabular}
\vspace{-5pt}
\caption{Multi-turn refinement results. We report the fix rate and the high-level score improvement for non-executable and executable code respectively.}
\label{tab:refinement_results}
\vspace{-5pt}
\end{table}

To evaluate the iterative debugging and refinement capabilities of our model, we conduct a two-turn experiment on ChartMimic and UniSVG. We first categorize the first-turn outputs into non-executable and executable groups to facilitate a granular analysis of different refinement behaviors. Specifically, we assess the repair rate of buggy code within the non-executable group and the degree of visual enhancement for the already executable group.
As shown in Table~\ref{tab:refinement_results}, for the non-executable group, our model demonstrates effective debugging skills, fixing 18.2\% of the buggy code on ChartMimic to make it runnable. For the executable group, the model focuses on visual refinement, further improving the high-level scores and narrowing the gap with the ground truth. These results show that with the refinement data, our model can effectively correct both functional errors and subtle visual details through multi-turn interactions.

\begin{table}[tbp]
\centering
\small
\setlength{\tabcolsep}{2pt}
\begin{tabular}{l|ccccc}
\toprule
\textbf{Model} & \textbf{ChartM.} & \textbf{UniSVG} & \textbf{D2C} & \textbf{I2S} & \textbf{ChemD.} \\
\midrule
Base & 50.3 & 63.3 & 85.5 & 34.1 & 16.4 \\
\quad + ViRL & 72.5\gap\upA{22.2} & 70.3\gap\upA{7.0} & 87.7\gap\upA{2.2} & 57.6\gap\upA{23.5} & 2.4\gap\downA{14.0} \\
\quad + SFT & 81.1\gap\upA{30.8} & 85.9\gap\upA{22.6} & 86.7\gap\upA{1.2} & 65.8\gap\upA{31.7} & 70.4\gap\upA{54.0} \\
\midrule
VinciCoder & 83.1 & 86.0 & 88.7 & 72.8 & 71.8 \\

\bottomrule
\end{tabular}
\vspace{-5pt}
\caption{Ablation study on training stages. We report the average of all metrics for each benchmark, except for UniSVG, which is evaluated using the overall score.}
\label{tab:training_stage_ablation}
\vspace{-10pt}
\end{table}

\subsection{Ablation Studies}
Furthermore, we conduct ablation studies on the VinciCoder-7B, focusing on training strategies, data scaling, and reward function designs.

\textbf{Training Strategy} We first evaluate the individual and combined contributions of the SFT and RL stages.
Table~\ref{tab:training_stage_ablation} confirms that our two-stage SFT-RL strategy achieves the best performance, validating the paradigm of using SFT to establish a strong policy followed by RL optimization. Notably, applying ViRL directly to the base model yields significant gains on ChartMimic and Design2Code, particularly in execution rate and high-level scores. This confirms that visual feedback effectively enhances both code correctness and perceptual fidelity. However, improvements on UniSVG are marginal, and performance even declines on ChemDraw, suggesting that the efficacy of ViRL is constrained by the base model's foundational capacity. However, as we only apply 42k data for RL and 1.6M for SFT, the results demonstrate the efficiency of our proposed ViRL strategy.


\begin{figure}[t]
    \centering
    \includegraphics[width=0.48\textwidth]{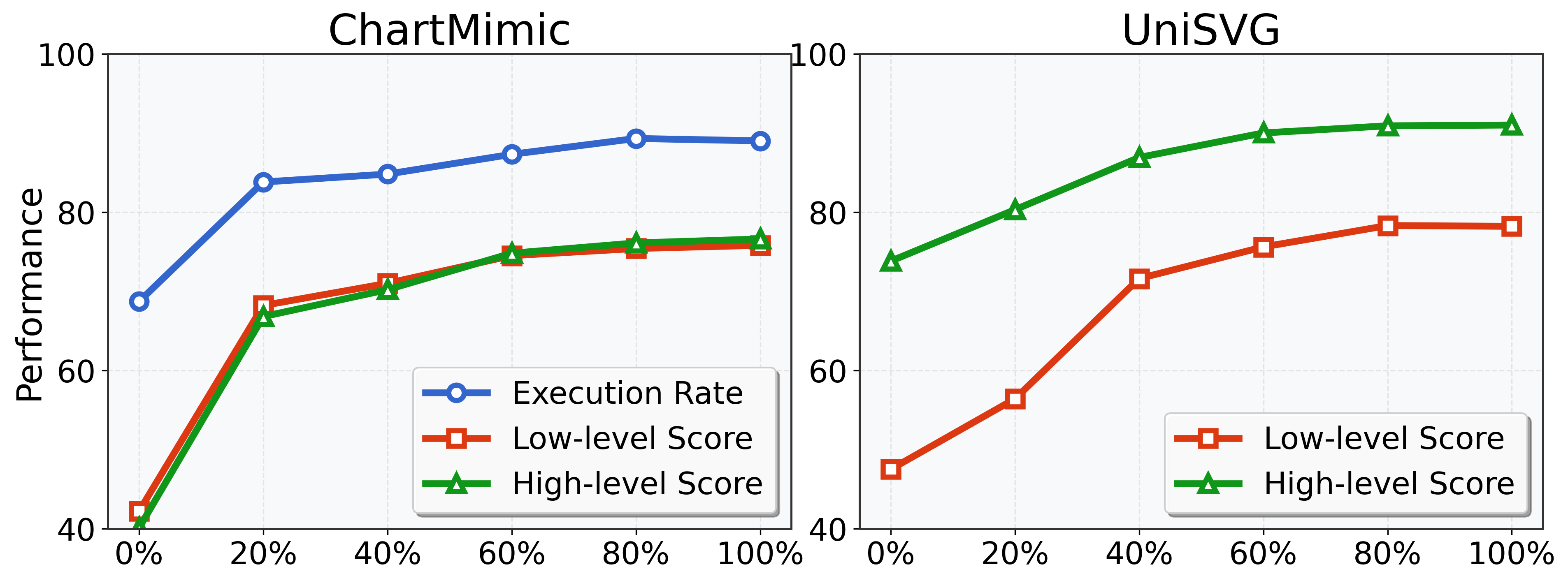}
    \vspace{-10pt}
    \caption{Ablation experiments about model performance under various SFT data scales.}
    \label{fig:ablation_sft_datascale}
    \vspace{-5pt}
\end{figure}

\begin{table}[tbp]
\centering
\small
\setlength{\tabcolsep}{1pt}
\begin{tabular}{l|ccccc}
\toprule
\textbf{Model} & \textbf{ChartM.} & \textbf{UniSVG} & \textbf{D2C} & \textbf{I2S} & \textbf{ChemD.} \\
\midrule
Full SFT & 81.1 & 85.9 & 86.7 & 65.8 & 70.4 \\
\quad + ViRL  & 83.1\gap\upB{2.0} & 86.0\gap\upB{0.1} & 88.7\gap\upB{2.0} & 72.8\gap\upB{7.0} & 71.8\gap\upB{1.4} \\
\midrule
20\% SFT & 72.9 & 70.8 &  86.0 & 49.0 &  56.1 \\
\quad + ViRL & 78.8\gap\upB{5.9} & 80.9\gap\upB{10.1} & 88.2\gap\upB{2.2} & 66.2\gap\upB{17.2} & 58.5\gap\upB{2.4} \\
\bottomrule
\end{tabular}
\vspace{-5pt}
\caption{Impact of SFT scale on ViRL performance. Relative improvements from ViRL are most pronounced at a 20\% SFT scale, underscoring its ability to enhance less-saturated models.}
\label{tab:sft_training_ablation}
\vspace{-10pt}
\end{table}

\begin{figure*}[t]
    \centering
    \includegraphics[width=0.96\textwidth]{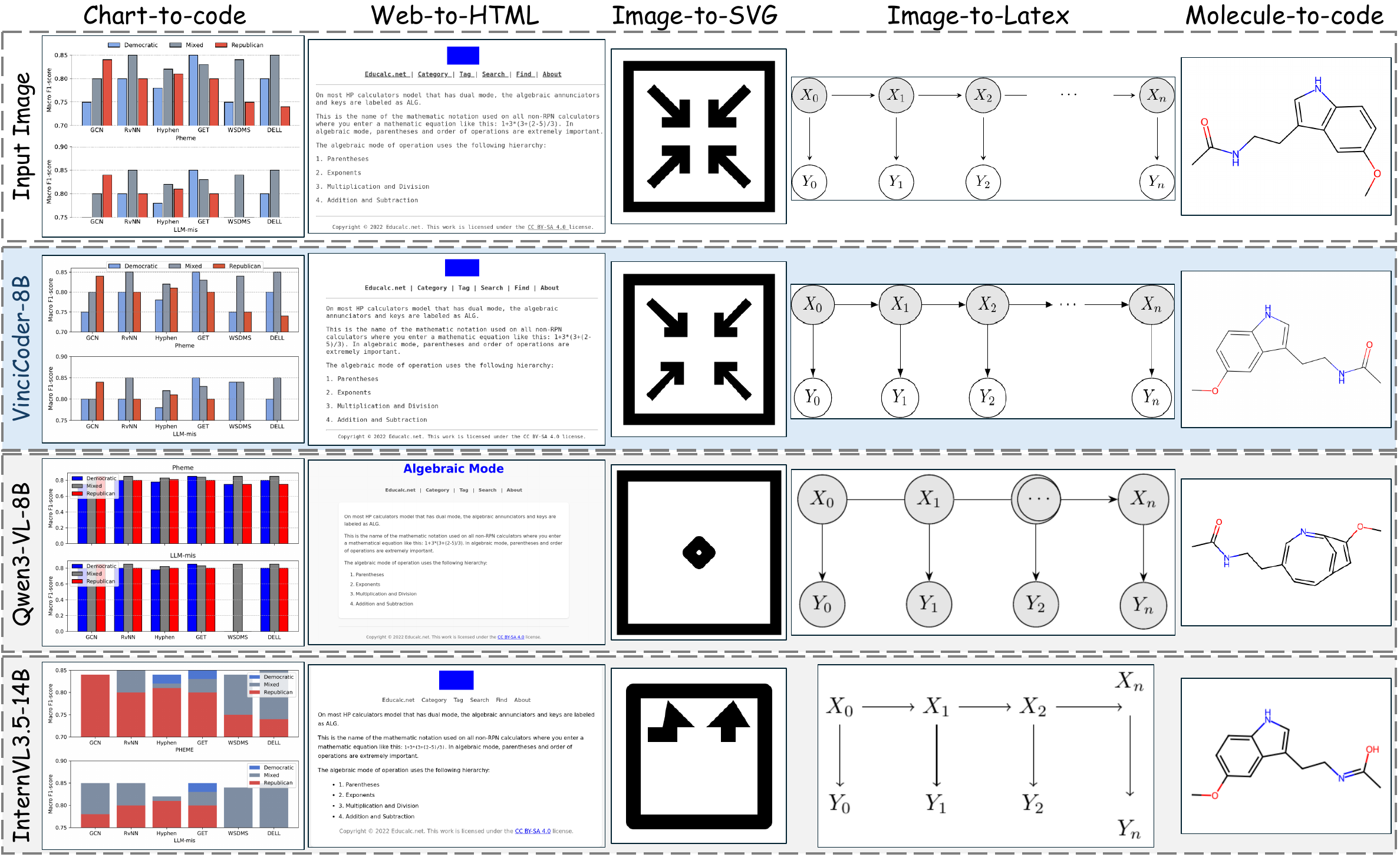}
    \vspace{-5pt}
    \caption{Showcases of images rendered by code generated by VinciCoder and other open-source VLMs.}
    \label{fig:showcase}
    \vspace{-5pt}
\end{figure*}

\textbf{Data Scaling.}
We further investigate how SFT data scale affects both immediate performance and the subsequent RL stage. As shown in Figure~\ref{fig:ablation_sft_datascale}, performance scales consistently with SFT dataset size before plateauing. This saturation suggests that while our large-scale SFT provides a robust foundation, further gains through supervised learning alone become marginal.

To isolate the contribution of our ViRL strategy, we employ it on the 20\% SFT model checkpoint. As detailed in Table~\ref{tab:sft_training_ablation}, ViRL yields a substantial performance leap in this setting. This confirms that the observed gains stem from the ViRL mechanism rather than SFT data scaling. While the 1.6M SFT provides a strong foundation, ViRL shows superior data efficiency in visual alignment, with its effectiveness being most pronounced before the base model reaches its SFT plateau. We provide the comprehensive results in Appendix~\ref{sec:full_results}.

\begin{table}[t]
\setlength{\tabcolsep}{3pt}
\small
\centering
\resizebox{0.48\textwidth}{!}{
\begin{tabular}{l|cc|cc|cc}
\toprule
\multirow{2}{*}{\textbf{Ablations}} & \multicolumn{2}{c}{\textbf{ChartMimic}} & \multicolumn{2}{c}{\textbf{UniSVG}} & \multicolumn{2}{c}{\textbf{Design2Code}} \\
\cmidrule{2-7}
& Low-L  & High-L & Low-L  & High-L & Low-L  & High-L \\
\midrule
\multicolumn{7}{l}{\textbf{Reward Formulations}} \\
\midrule
Edit Dist.2 & 77.1 & 76.7 & 75.9 & 88.5 & 87.1 & 87.4 \\
SSIM & 75.7 & 75.3 & 77.2 & 89.8 & 84.5 & 85.6 \\
InternViT & 77.9 & 78.1 & 77.1 & 90.6 & 86.3 & 87.9\\
\midrule
\multicolumn{7}{l}{\textbf{Coarse-to-fine ViRL}} \\
\midrule
w/o Coarse & 77.8 & 78.9 & 76.9 & 91.8 & 87.4 & 88.3\\
w/o Fine & 77.0 & 78.0 & 77.1 & 91.8 & 86.9 & 87.6\\
w/o Align & 78.2 & 79.4 & 76.2 & 90.5 & 87.5 & 88.3\\
\midrule
VinciCoder  & 78.3 & 79.8 & 77.0 & 92.0 & 88.2 & 89.1 \\
\bottomrule
\end{tabular}}
\vspace{-5pt}
\caption{Ablation studies about the reward function.}
\label{tab:ablation_reward}
\vspace{-10pt}
\end{table}

\textbf{Reward Functions.} Last but not least, we conduct ablation studies to validate our coarse-to-fine visual reward design. We evaluate our approach against three baselines: a textual reward based on edit distance, a low-level visual reward based on SSIM, and a semantic-level visual reward based on InternViT-300M (with matched parameters with DINO-L). Furthermore, we isolate the impact of individual components within our coarse-to-fine strategy to verify their specific contributions.
As shown in Table~\ref{tab:ablation_reward}, our coarse-to-fine reward achieves the most balanced performance. Analysis indicates that self-supervised DINOv2 excels at capturing fine-grained features essential for visual fidelity, whereas image-text aligned ViTs prioritize global semantic alignment. Furthermore, the coarse-to-fine strategy is critical for high-resolution benchmarks like ChartMimic and Design2Code, where omitting either component leads to significant degradation. Conversely, performance on UniSVG remains stable, as its lower resolution allows our method to simplify into a global comparison, rendering the coarse and fine-grained rewards functionally equivalent.

We provide qualitative showcases in Figure~\ref{fig:showcase}. Also, additional results and detailed analysis such as failure modes, are illustrated in Appendix~\ref{sec:further_results}.

\section{Conclusion}
In this paper, we present VinciCoder, a unified multimodal code generation framework optimized via a two-stage SFT-ViRL strategy. Following large-scale SFT on 1.6M multi-domain samples, we introduce ViRL, a reinforcement learning paradigm featuring a coarse-to-fine reward mechanism that aligns code generation with visual perceptual signals. Extensive benchmarks demonstrate that VinciCoder significantly outperforms existing open-source models. Our ablation studies further validate the efficacy of the ViRL paradigm and its ability to achieve high-fidelity visual alignment.

\section*{Limitations}
Despite improvements in visual quality and code performance, VinciCoder has some limitations. First, visual feedback helps the model organize layout and style, but it cannot teach the model new domain knowledge that is missing from the base policy. For example, if the model does not understand complex chemical structures in ChemDraw, the visual reward might cause it to take shortcuts. In these cases, the model creates basic shapes that look like the target but are logically incorrect. Second, the rendering process and the need to generate many samples for reinforcement learning require a lot of computing power. This makes it difficult to scale the model for very long code files or high resolution images. Finally, the reward system is based on static images. It does not yet consider interactive or moving parts in web based code like HTML and JavaScript.

\section*{Ethic Statement}
The development of VinciCoder follows ethical guidelines for data usage and model deployment. Our project uses established open source models and libraries to ensure that our research is transparent and easy to reproduce. All training data includes 1.6M SFT samples and 42k RL samples collected from public sources according to their original licenses. We used strict filters to remove personally identifiable information from the training process. Although VinciCoder improves multimodal code generation, it is designed for research and education. Users must know that automated code generation can produce incorrect results. Therefore, we recommend human review for safety critical software. We do not support the use of this model to create malicious code or violate intellectual property rights. By using and supporting the open source community, we hope to encourage responsible progress in vision language modeling.
\bibliography{custom}

@article{li2025unisvg,
  title={UniSVG: A Unified Dataset for Vector Graphic Understanding and Generation with Multimodal Large Language Models},
  author={Li, Jinke and Yu, Jiarui and Wei, Chenxing and Dong, Hande and Lin, Qiang and Yang, Liangjing and Wang, Zhicai and Hao, Yanbin},
  journal={arXiv preprint arXiv:2508.07766},
  year={2025}
}

@article{tan2025reason,
  title={Reason-rft: Reinforcement fine-tuning for visual reasoning},
  author={Tan, Huajie and Ji, Yuheng and Hao, Xiaoshuai and Lin, Minglan and Wang, Pengwei and Wang, Zhongyuan and Zhang, Shanghang},
  journal={arXiv preprint arXiv:2503.20752},
  year={2025}
}

@article{liu2025seg,
  title={Seg-zero: Reasoning-chain guided segmentation via cognitive reinforcement},
  author={Liu, Yuqi and Peng, Bohao and Zhong, Zhisheng and Yue, Zihao and Lu, Fanbin and Yu, Bei and Jia, Jiaya},
  journal={arXiv preprint arXiv:2503.06520},
  year={2025}
}

@article{liu2025visual,
  title={Visual-rft: Visual reinforcement fine-tuning},
  author={Liu, Ziyu and Sun, Zeyi and Zang, Yuhang and Dong, Xiaoyi and Cao, Yuhang and Duan, Haodong and Lin, Dahua and Wang, Jiaqi},
  journal={arXiv preprint arXiv:2503.01785},
  year={2025}
}

@article{zhang2025artifactsbench,
  title={Artifactsbench: Bridging the visual-interactive gap in llm code generation evaluation},
  author={Zhang, Chenchen and Li, Yuhang and Xu, Can and Liu, Jiaheng and Liu, Ao and Zhou, Changzhi and Deng, Ken and Wu, Dengpeng and Huang, Guanhua and Li, Kejiao and others},
  journal={arXiv preprint arXiv:2507.04952},
  year={2025}
}

@article{shen2025vlm,
  title={Vlm-r1: A stable and generalizable r1-style large vision-language model},
  author={Shen, Haozhan and Liu, Peng and Li, Jingcheng and Fang, Chunxin and Ma, Yibo and Liao, Jiajia and Shen, Qiaoli and Zhang, Zilun and Zhao, Kangjia and Zhang, Qianqian and others},
  journal={arXiv preprint arXiv:2504.07615},
  year={2025}
}

@article{chen2025svgenius,
  title={SVGenius: Benchmarking LLMs in SVG Understanding, Editing and Generation},
  author={Chen, Siqi and Dong, Xinyu and Xu, Haolei and Wu, Xingyu and Tang, Fei and Zhang, Hang and Yan, Yuchen and Wu, Linjuan and Zhang, Wenqi and Hou, Guiyang and others},
  journal={arXiv preprint arXiv:2506.03139},
  year={2025}
}

@article{yang2024chartmimic,
  title={Chartmimic: Evaluating lmm's cross-modal reasoning capability via chart-to-code generation},
  author={Yang, Cheng and Shi, Chufan and Liu, Yaxin and Shui, Bo and Wang, Junjie and Jing, Mohan and Xu, Linran and Zhu, Xinyu and Li, Siheng and Zhang, Yuxiang and others},
  journal={arXiv preprint arXiv:2406.09961},
  year={2024}
}

@article{wu2024plot2code,
  title={Plot2code: A comprehensive benchmark for evaluating multi-modal large language models in code generation from scientific plots},
  author={Wu, Chengyue and Ge, Yixiao and Guo, Qiushan and Wang, Jiahao and Liang, Zhixuan and Lu, Zeyu and Shan, Ying and Luo, Ping},
  journal={arXiv preprint arXiv:2405.07990},
  year={2024}
}

@article{li2025opusanimation,
  title={OpusAnimation: Code-Based Dynamic Chart Generation},
  author={Li, Bozheng and Yang, Miao and Chen, Zhenhan and Cao, Jiawang and Liu, Mushui and Lu, Yi and Wu, Yongliang and Zhang, Bin and Ji, Yangguang and Tang, Licheng and others},
  journal={arXiv preprint arXiv:2510.03341},
  year={2025}
}

@inproceedings{rodriguez2025starvector,
  title={Starvector: Generating scalable vector graphics code from images and text},
  author={Rodriguez, Juan A and Puri, Abhay and Agarwal, Shubham and Laradji, Issam H and Rodriguez, Pau and Rajeswar, Sai and Vazquez, David and Pal, Christopher and Pedersoli, Marco},
  booktitle={Proceedings of the Computer Vision and Pattern Recognition Conference},
  pages={16175--16186},
  year={2025}
}

@article{wu2025mllm,
  title={MLLM-Based UI2Code Automation Guided by UI Layout Information},
  author={Wu, Fan and Gao, Cuiyun and Li, Shuqing and Wen, Xin-Cheng and Liao, Qing},
  journal={Proceedings of the ACM on Software Engineering},
  volume={2},
  number={ISSTA},
  pages={1123--1145},
  year={2025},
  publisher={ACM New York, NY, USA}
}

@inproceedings{gui2025webcode2m,
  title={Webcode2m: A real-world dataset for code generation from webpage designs},
  author={Gui, Yi and Li, Zhen and Wan, Yao and Shi, Yemin and Zhang, Hongyu and Chen, Bohua and Su, Yi and Chen, Dongping and Wu, Siyuan and Zhou, Xing and others},
  booktitle={Proceedings of the ACM on Web Conference 2025},
  pages={1834--1845},
  year={2025}
}

@article{shao2024deepseekmath,
  title={Deepseekmath: Pushing the limits of mathematical reasoning in open language models},
  author={Shao, Zhihong and Wang, Peiyi and Zhu, Qihao and Xu, Runxin and Song, Junxiao and Bi, Xiao and Zhang, Haowei and Zhang, Mingchuan and Li, YK and Wu, Yang and others},
  journal={arXiv preprint arXiv:2402.03300},
  year={2024}
}

@article{yun2024web2code,
  title={Web2code: A large-scale webpage-to-code dataset and evaluation framework for multimodal llms},
  author={Yun, Sukmin and Thushara, Rusiru and Bhat, Mohammad and Wang, Yongxin and Deng, Mingkai and Wang, Jinhong and Tao, Tianhua and Li, Junbo and Li, Haonan and Nakov, Preslav and others},
  journal={Advances in neural information processing systems},
  volume={37},
  pages={112134--112157},
  year={2024}
}

@article{guo2025deepseek,
  title={Deepseek-r1 incentivizes reasoning in llms through reinforcement learning},
  author={Guo, Daya and Yang, Dejian and Zhang, Haowei and Song, Junxiao and Wang, Peiyi and Zhu, Qihao and Xu, Runxin and Zhang, Ruoyu and Ma, Shirong and Bi, Xiao and others},
  journal={Nature},
  volume={645},
  number={8081},
  pages={633--638},
  year={2025},
  publisher={Nature Publishing Group UK London}
}

@article{roberts2024image2struct,
  title={Image2struct: Benchmarking structure extraction for vision-language models},
  author={Roberts, Josselin S and Lee, Tony and Wong, Chi H and Yasunaga, Michihiro and Mai, Yifan and Liang, Percy},
  journal={Advances in Neural Information Processing Systems},
  volume={37},
  pages={115058--115097},
  year={2024}
}

@article{chen2025breaking,
  title={Breaking the sft plateau: Multimodal structured reinforcement learning for chart-to-code generation},
  author={Chen, Lei and Zhao, Xuanle and Zeng, Zhixiong and Huang, Jing and Zheng, Liming and Zhong, Yufeng and Ma, Lin},
  journal={arXiv preprint arXiv:2508.13587},
  year={2025}
}

@article{si2024design2code,
  title={Design2code: Benchmarking multimodal code generation for automated front-end engineering},
  author={Si, Chenglei and Zhang, Yanzhe and Li, Ryan and Yang, Zhengyuan and Liu, Ruibo and Yang, Diyi},
  journal={arXiv preprint arXiv:2403.03163},
  year={2024}
}

@article{yang2025chartm,
  title={ChartM$^3$: Benchmarking Chart Editing with Multimodal Instructions},
  author={Yang, Donglu and Zhang, Liang and Yue, Zihao and Chen, Liangyu and Xu, Yichen and Wang, Wenxuan and Jin, Qin},
  journal={arXiv preprint arXiv:2507.21167},
  year={2025}
}

@article{yang2025scaling,
  title={Scaling text-rich image understanding via code-guided synthetic multimodal data generation},
  author={Yang, Yue and Patel, Ajay and Deitke, Matt and Gupta, Tanmay and Weihs, Luca and Head, Andrew and Yatskar, Mark and Callison-Burch, Chris and Krishna, Ranjay and Kembhavi, Aniruddha and others},
  journal={arXiv preprint arXiv:2502.14846},
  year={2025}
}

@article{jiang2025viscodex,
  title={VisCodex: Unified Multimodal Code Generation via Merging Vision and Coding Models},
  author={Jiang, Lingjie and Huang, Shaohan and Wu, Xun and Li, Yixia and Zhang, Dongdong and Wei, Furu},
  journal={arXiv preprint arXiv:2508.09945},
  year={2025}
}

@article{wang2025mathcoder,
  title={MathCoder-VL: Bridging Vision and Code for Enhanced Multimodal Mathematical Reasoning},
  author={Wang, Ke and Pan, Junting and Wei, Linda and Zhou, Aojun and Shi, Weikang and Lu, Zimu and Xiao, Han and Yang, Yunqiao and Ren, Houxing and Zhan, Mingjie and others},
  journal={arXiv preprint arXiv:2505.10557},
  year={2025}
}

@article{belouadi2025tikzero,
  title={TikZero: Zero-Shot Text-Guided Graphics Program Synthesis},
  author={Belouadi, Jonas and Ilg, Eddy and Keuper, Margret and Tanaka, Hideki and Utiyama, Masao and Dabre, Raj and Eger, Steffen and Ponzetto, Simone Paolo},
  journal={arXiv preprint arXiv:2503.11509},
  year={2025}
}

@article{yang2025omnisvg,
  title={Omnisvg: A unified scalable vector graphics generation model},
  author={Yang, Yiying and Cheng, Wei and Chen, Sijin and Zeng, Xianfang and Yin, Fukun and Zhang, Jiaxu and Wang, Liao and Yu, Gang and Ma, Xingjun and Jiang, Yu-Gang},
  journal={arXiv preprint arXiv:2504.06263},
  year={2025}
}

@article{belouadi2024detikzify,
  title={Detikzify: Synthesizing graphics programs for scientific figures and sketches with tikz},
  author={Belouadi, Jonas and Ponzetto, Simone and Eger, Steffen},
  journal={Advances in Neural Information Processing Systems},
  volume={37},
  pages={85074--85108},
  year={2024}
}

@article{tan2025chartmaster,
  title={Chartmaster: Advancing chart-to-code generation with real-world charts and chart similarity reinforcement learning},
  author={Tan, Wentao and Cao, Qiong and Xue, Chao and Zhan, Yibing and Ding, Changxing and He, Xiaodong},
  journal={arXiv preprint arXiv:2508.17608},
  year={2025}
}

@article{yang2025qwen3,
  title={Qwen3 technical report},
  author={Yang, An and Li, Anfeng and Yang, Baosong and Zhang, Beichen and Hui, Binyuan and Zheng, Bo and Yu, Bowen and Gao, Chang and Huang, Chengen and Lv, Chenxu and others},
  journal={arXiv preprint arXiv:2505.09388},
  year={2025}
}

@article{comanici2025gemini,
  title={Gemini 2.5: Pushing the frontier with advanced reasoning, multimodality, long context, and next generation agentic capabilities},
  author={Comanici, Gheorghe and Bieber, Eric and Schaekermann, Mike and Pasupat, Ice and Sachdeva, Noveen and Dhillon, Inderjit and Blistein, Marcel and Ram, Ori and Zhang, Dan and Rosen, Evan and others},
  journal={arXiv preprint arXiv:2507.06261},
  year={2025}
}

@article{oquab2023dinov2,
  title={Dinov2: Learning robust visual features without supervision},
  author={Oquab, Maxime and Darcet, Timoth{\'e}e and Moutakanni, Th{\'e}o and Vo, Huy and Szafraniec, Marc and Khalidov, Vasil and Fernandez, Pierre and Haziza, Daniel and Massa, Francisco and El-Nouby, Alaaeldin and others},
  journal={arXiv preprint arXiv:2304.07193},
  year={2023}
}

@article{zhang2025boosting,
  title={Boosting Chart-to-Code Generation in MLLM via Dual Preference-Guided Refinement},
  author={Zhang, Zhihan and Cao, Yixin and Liao, Lizi},
  journal={arXiv preprint arXiv:2504.02906},
  year={2025}
}

@article{rodriguez2025rendering,
  title={Rendering-Aware Reinforcement Learning for Vector Graphics Generation},
  author={Rodriguez, Juan A and Zhang, Haotian and Puri, Abhay and Feizi, Aarash and Pramanik, Rishav and Wichmann, Pascal and Mondal, Arnab and Samsami, Mohammad Reza and Awal, Rabiul and Taslakian, Perouz and others},
  journal={arXiv preprint arXiv:2505.20793},
  year={2025}
}

@article{zhu2025internvl3,
  title={Internvl3: Exploring advanced training and test-time recipes for open-source multimodal models},
  author={Zhu, Jinguo and Wang, Weiyun and Chen, Zhe and Liu, Zhaoyang and Ye, Shenglong and Gu, Lixin and Tian, Hao and Duan, Yuchen and Su, Weijie and Shao, Jie and others},
  journal={arXiv preprint arXiv:2504.10479},
  year={2025}
}

@article{wang2025internvl3,
  title={Internvl3. 5: Advancing open-source multimodal models in versatility, reasoning, and efficiency},
  author={Wang, Weiyun and Gao, Zhangwei and Gu, Lixin and Pu, Hengjun and Cui, Long and Wei, Xingguang and Liu, Zhaoyang and Jing, Linglin and Ye, Shenglong and Shao, Jie and others},
  journal={arXiv preprint arXiv:2508.18265},
  year={2025}
}

@article{landrum2013rdkit,
  title={Rdkit documentation},
  author={Landrum, Greg},
  journal={Release},
  volume={1},
  number={1-79},
  pages={4},
  year={2013}
}

@article{wan2024automatically,
  title={Automatically generating UI code from screenshot: A divide-and-conquer-based approach},
  author={Wan, Yuxuan and Wang, Chaozheng and Dong, Yi and Wang, Wenxuan and Li, Shuqing and Huo, Yintong and Lyu, Michael R},
  journal={arXiv preprint arXiv:2406.16386},
  year={2024}
}

@article{jiang2025screencoder,
  title={Screencoder: Advancing visual-to-code generation for front-end automation via modular multimodal agents},
  author={Jiang, Yilei and Zheng, Yaozhi and Wan, Yuxuan and Han, Jiaming and Wang, Qunzhong and Lyu, Michael R and Yue, Xiangyu},
  journal={arXiv preprint arXiv:2507.22827},
  year={2025}
}

@article{schulman2017proximal,
  title={Proximal policy optimization algorithms},
  author={Schulman, John and Wolski, Filip and Dhariwal, Prafulla and Radford, Alec and Klimov, Oleg},
  journal={arXiv preprint arXiv:1707.06347},
  year={2017}
}

@article{sun2025januscoder,
  title={JanusCoder: Towards a Foundational Visual-Programmatic Interface for Code Intelligence},
  author={Sun, Qiushi and Gong, Jingyang and Liu, Yang and Chen, Qiaosheng and Li, Lei and Chen, Kai and Guo, Qipeng and Kao, Ben and Yuan, Fei},
  journal={arXiv preprint arXiv:2510.23538},
  year={2025}
}

@misc{anthropic_claude_4_5_sonnet,
  author       = {{Anthropic}},
  title        = {Introducing Claude Sonnet 4.5},
  howpublished = {Web Page},
  year         = {2025},
  url          = {https://www.anthropic.com/news/claude-sonnet-4-5}
}

@misc{openai_gpt_5,
  author       = {{OpenAI}},
  title        = {Introducing GPT-5},
  howpublished = {Web Page},
  year         = {2025},
  url          = {https://openai.com/zh-Hans-CN/index/introducing-gpt-5/}
}

@article{bai2025qwen2,
  title={Qwen2.5-vl technical report},
  author={Bai, Shuai and Chen, Keqin and Liu, Xuejing and Wang, Jialin and Ge, Wenbin and Song, Sibo and Dang, Kai and Wang, Peng and Wang, Shijie and Tang, Jun and others},
  journal={arXiv preprint arXiv:2502.13923},
  year={2025}
}

@misc{IndigoToolkit,
  author       = {{EPAM Systems}},
  title        = {Indigo Toolkit},
  note         = {Accessed: 2025-10-18},
  year         = {2025}
}

@article{zhao2025chartcoder,
  title={Chartcoder: Advancing multimodal large language model for chart-to-code generation},
  author={Zhao, Xuanle and Luo, Xianzhen and Shi, Qi and Chen, Chi and Wang, Shuo and Liu, Zhiyuan and Sun, Maosong},
  journal={arXiv preprint arXiv:2501.06598},
  year={2025}
}

@article{huang2025vision,
  title={Vision-r1: Incentivizing reasoning capability in multimodal large language models},
  author={Huang, Wenxuan and Jia, Bohan and Zhai, Zijie and Cao, Shaosheng and Ye, Zheyu and Zhao, Fei and Xu, Zhe and Hu, Yao and Lin, Shaohui},
  journal={arXiv preprint arXiv:2503.06749},
  year={2025}
}

@article{chen2025learning,
  title={Learning only with images: Visual reinforcement learning with reasoning, rendering, and visual feedback},
  author={Chen, Yang and Shen, Yufan and Huang, Wenxuan and Zhou, Sheng and Lin, Qunshu and Cai, Xinyu and Yu, Zhi and Bu, Jiajun and Shi, Botian and Qiao, Yu},
  journal={arXiv preprint arXiv:2507.20766},
  year={2025}
}

@article{xiao2024interaction2code,
  title={Interaction2code: How far are we from automatic interactive webpage generation?},
  author={Xiao, Jingyu and Wan, Yuxuan and Huo, Yintong and Xu, Zhiyao and Lyu, Michael R},
  journal={arXiv e-prints},
  pages={arXiv--2411},
  year={2024}
}

@article{li2024sketch2code,
  title={Sketch2code: Evaluating vision-language models for interactive web design prototyping},
  author={Li, Ryan and Zhang, Yanzhe and Yang, Diyi},
  journal={arXiv preprint arXiv:2410.16232},
  year={2024}
}

@article{yang2025r1,
  title={R1-onevision: Advancing generalized multimodal reasoning through cross-modal formalization},
  author={Yang, Yi and He, Xiaoxuan and Pan, Hongkun and Jiang, Xiyan and Deng, Yan and Yang, Xingtao and Lu, Haoyu and Yin, Dacheng and Rao, Fengyun and Zhu, Minfeng and others},
  journal={arXiv preprint arXiv:2503.10615},
  year={2025}
}

@article{chen2025chart,
  title={Chart-r1: Chain-of-thought supervision and reinforcement for advanced chart reasoner},
  author={Chen, Lei and Zhao, Xuanle and Zeng, Zhixiong and Huang, Jing and Zhong, Yufeng and Ma, Lin},
  journal={arXiv preprint arXiv:2507.15509},
  year={2025}
}

@article{zhao2025chartedit,
  title={ChartEdit: How Far Are MLLMs From Automating Chart Analysis? Evaluating MLLMs' Capability via Chart Editing},
  author={Zhao, Xuanle and Liu, Xuexin and Yang, Haoyue and Luo, Xianzhen and Zeng, Fanhu and Li, Jianling and Shi, Qi and Chen, Chi},
  journal={arXiv preprint arXiv:2505.11935},
  year={2025}
}

@article{zhou2025gui,
  title={Gui-g1: Understanding r1-zero-like training for visual grounding in gui agents},
  author={Zhou, Yuqi and Dai, Sunhao and Wang, Shuai and Zhou, Kaiwen and Jia, Qinglin and Xu, Jun},
  journal={arXiv preprint arXiv:2505.15810},
  year={2025}
}

@misc{bai2025qwen3vltechnicalreport,
      title={Qwen3-VL Technical Report}, 
      author={Shuai Bai and Yuxuan Cai and Ruizhe Chen and Keqin Chen and Xionghui Chen and Zesen Cheng and Lianghao Deng and Wei Ding and Chang Gao and Chunjiang Ge and Wenbin Ge and Zhifang Guo and Qidong Huang and Jie Huang and Fei Huang and Binyuan Hui and Shutong Jiang and Zhaohai Li and Mingsheng Li and Mei Li and Kaixin Li and Zicheng Lin and Junyang Lin and Xuejing Liu and Jiawei Liu and Chenglong Liu and Yang Liu and Dayiheng Liu and Shixuan Liu and Dunjie Lu and Ruilin Luo and Chenxu Lv and Rui Men and Lingchen Meng and Xuancheng Ren and Xingzhang Ren and Sibo Song and Yuchong Sun and Jun Tang and Jianhong Tu and Jianqiang Wan and Peng Wang and Pengfei Wang and Qiuyue Wang and Yuxuan Wang and Tianbao Xie and Yiheng Xu and Haiyang Xu and Jin Xu and Zhibo Yang and Mingkun Yang and Jianxin Yang and An Yang and Bowen Yu and Fei Zhang and Hang Zhang and Xi Zhang and Bo Zheng and Humen Zhong and Jingren Zhou and Fan Zhou and Jing Zhou and Yuanzhi Zhu and Ke Zhu},
      year={2025},
      eprint={2511.21631},
      archivePrefix={arXiv},
      primaryClass={cs.CV},
      url={https://arxiv.org/abs/2511.21631}, 
}

\appendix

\clearpage

\section{Benchmark Details}
Our evaluation methodology for the ChartMimic, UniSVG, and Design2Code benchmarks follows the official implementations provided in their respective GitHub repositories. Regarding Image2Struct, we observed a minor inconsistency between the metrics in the published paper and the official code. To ensure replicability, we follow the implementation in the GitHub repository. We introduce the our calculation of EMS in Section~\ref{sec:ms}. For the ChemDraw benchmark, we evaluate performance using the execution rate and Tanimoto similarity, two metrics we introduce in Section~\ref{sec:chemdraw}.

\subsection{Earth Mover's Similarity (EMS)}\label{sec:ms}
The Earth Mover's Similarity (EMS) is an efficient, patch-based metric derived from the Earth Mover's Distance (EMD). It quantifies similarity through a two-level process that compares the arrangement of image patches at a global level and the pixel distributions within those patches at a local level.

The calculation begins by converting the reference image $x$ and the generated image $\hat{x}$ to grayscale. The most frequent pixel value in the reference image is identified and designated as the background color, $v_{bg}$, to focus the analysis on foreground content.

Next, both images are partitioned into a grid of $K$ patches, $\{P_0, \dots, P_{K-1}\}$. A global signature, $S_{global} = \{(w_i, P_i, c_i)\}_{i=0}^{K-1}$, is created for each image. Each element in the signature represents a patch $P_i$ by its content, its normalized center coordinate $c_i$, and a weight $w_i$. To prioritize meaningful content, this weight $w_i$ is set significantly higher for patches containing non-background pixels.

The dissimilarity between the images is calculated using a hierarchical cost matrix $C_p$, where each element $C_p[i, j]$ defines the cost of matching patch $P_i$ from image $x$ to patch $P_j$ from image $\hat{x}$. This cost aggregates two factors: the dissimilarity within the patches and the spatial distance between them:
\begin{equation}
C_p[i, j] = \text{EMD}_{intra}(P_i, P_j) + \lambda \cdot ||c_i - c_j||_1
\end{equation}

\begin{algorithm}[t]
\caption{Tanimoto Similarity Calculation}
\label{alg:tanimoto}
\small
\begin{algorithmic}[1]
\Function{CalculateTanimoto}{pred\_smiles, gt\_smiles}
    \\
    \Comment{Step 1: Convert SMILES strings to RDKit molecule.}
    \State $\textit{pred\_mol} \gets \text{RDKit.MolFromSmiles}(\textit{pred\_smiles})$
    \State $\textit{gt\_mol} \gets \text{RDKit.MolFromSmiles}(\textit{gt\_smiles})$
    \
    \Comment{Step 2: Handle invalid SMILES. If either is invalid, similarity is 0.}
    \If{$\textit{pred\_mol}$ is \textbf{null} \textbf{or} $\textit{gt\_mol}$ is \textbf{null}}
        \State \Return $0.0$
    \EndIf
    \\
    \Comment{Step 3: Generate molecular fingerprints (e.g., Morgan fingerprints).}
    \State $\textit{pred\_fp} \gets \text{GetMorganFP}(\textit{pred\_mol}, \text{radius}=2)$
    \State $\textit{gt\_fp} \gets \text{GetMorganFP}(\textit{gt\_mol}, \text{radius}=2)$
    \\
    \Comment{Step 4: Calculate Tanimoto similarity between the two fingerprints.}
    \State $\textit{similarity} \gets \text{TanimotoSimilarity}(\textit{pred\_fp}, \textit{gt\_fp})$
    
    \State \Return $\textit{similarity}$
\EndFunction
\end{algorithmic}
\end{algorithm}

Here, $\text{EMD}_{intra}$ is the classic EMD computed on the pixel values and their local coordinates within each patch. The term $||c_i - c_j||_1$ is the $L_1$ (Manhattan) distance between the patch centers, scaled by a factor $\lambda$:
\begin{equation}
\lambda = \frac{\sqrt{r \times s}}{W + H}
\end{equation}
where $(r, s)$ are the patch dimensions and $(W, H)$ are the image dimensions.

The total dissimilarity score, $\text{EMD}_{block}(x, \hat{x})$, is obtained by solving the transportation problem defined by the cost matrix $C_p$. This score is then normalized to produce the final EMS value, which lies in the range $[0, 1]$. Normalization is achieved by dividing by the dissimilarity between the reference image $x$ and a constant image $x_{const}$ (either black or white, whichever is more dissimilar):
\begin{equation}
\text{EMS}(x, \hat{x}) = \max \left(0, 1 - \frac{\text{EMD}_{block}(x, \hat{x})}{\text{EMD}_{block}(x, x_{const})} \right)
\end{equation}
An EMS score of 1 indicates identical images, while a score of 0 indicates maximum dissimilarity.

\begin{table*}[t]
\setlength{\tabcolsep}{2pt}
\small
\centering
\resizebox{\textwidth}{!}{
\begin{tabular}{l|ccc|ccc|cc|cc|cc}
\toprule
\multirow{2}{*}{Model} & \multicolumn{3}{c|}{\textbf{ChartMimic-direct-v2}} & \multicolumn{3}{c|}{\textbf{UniSVG-ISVGEN}} & \multicolumn{2}{c|}{\textbf{Design2Code}} & 
\multicolumn{2}{c|}{\textbf{Image2Struct}} & \multicolumn{2}{c}{\textbf{ChemDraw}} \\
\cmidrule{2-13}
   & Exec.Rate & Low-L  & High-L & Low-L  & High-L  & Score & Low-L & High-L & Ren.Succ. & EMS & Exec.Rate & Tani.Sim.\\ 
\midrule
\multicolumn{13}{l}{\textbf{VinciCoder-7B Variants}} \\
\midrule
Full SFT & 89.0 & 75.8 & 78.6 & 78.2 & 91.0 & 85.9 & 86.5 & 87.0 & 77.0 & 54.6  & 85.9 & 54.9 \\
SFT-ViRL  & 91.2 & 78.3 & 79.8 & 77.0 & 92.0 & 86.0 & 88.2 & 89.1  & 84.7 & 60.9 & 87.5 & 56.0 \\
20\%SFT  & 83.8 & 68.2 & 66.8 & 56.4 & 80.3 & 70.8 & 85.6 & 86.4 & 57.0 & 40.9 &  75.8  & 36.4\\
20\%SFT-ViRL  & 88.3 & 73.4  & 74.7 & 69.4 & 88.5 & 80.9 & 87.7 & 88.7 & 77.7 & 54.8 & 78.9 & 38.0\\
ViRL Only & 90.2 & 58.6 & 68.8 & 55.2 & 80.3 & 70.3 & 86.8 & 88.6 & 64.3 & 50.9 & 4.1 & 0.7 \\
\midrule
\multicolumn{13}{l}{\textbf{VinciCoder-8B Variants}} \\
\midrule
Full SFT & 88.3 & 75.6 & 78.9 & 78.4 & 93.7 &  87.6 & 86.8 & 87.9 & 72.3 & 50.7 & 85.9 & 59.3\\
SFT-ViRL & 91.6 & 78.9 & 80.6 & 77.1 & 94.1 & 87.3 & 88.4 & 89.3 & 77.3 & 57.8 & 88.3 & 62.6\\
\bottomrule 
\end{tabular}
}
\vspace{-5pt}
\caption{Comprehensive results for VinciCoder variants across different benchmarks.}
\label{tab:vincicoder_variant}
\vspace{-10pt}
\end{table*}

\subsection{ChemDraw Metrics}\label{sec:chemdraw}
Performance on the molecule-to-code task is assessed using two key metrics. The first, execution rate, serves as a primary filter for syntactic correctness, evaluating whether the VLM produces chemically valid code and SMILES strings. The second, Tanimoto similarity, measures the chemical fidelity of the valid generations. This metric utilizes the RDKit library to compare the molecular fingerprints of the generated structure against the ground-truth structure. A detailed description of the calculation in Algorithm~\ref{alg:tanimoto}.

The formula for Tanimoto similarity between two fingerprints (bit vectors) $A$ and $B$ is:
$$
T(A, B) = \frac{|A \cap B|}{|A \cup B|} = \frac{c}{a + b - c}
$$
where:
\begin{itemize}
    \item $a$ is the number of bits set in fingerprint $A$.
    \item $b$ is the number of bits set in fingerprint $B$.
    \item $c$ is the number of bits set in both $A$ and $B$ (the intersection).
\end{itemize}

\section{Further Experiment Results}\label{sec:further_results}
\subsection{Full Results of VinciCoder Variants}\label{sec:full_results}
We provide full evaluation results for VinciCoder variants to show the impact of different training stages and model scales. As shown in Table~\ref{tab:vincicoder_variant}, we report performance for both 7B and 8B versions at the SFT checkpoint (SFT Only) and the checkpoint after SFT and ViRL training (SFT-ViRL), which correspond to VinciCoder-7B and -8B in Table~\ref{tab:main_results}. These results demonstrate how the ViRL strategy improves the model performance after the initial SFT stage.

In addition to the main models, we provide the full ablation results of the 7B variant. This includes the 20 percent SFT model (20\%SFT) and its corresponding RL version (20\%SFT-ViRL) as well as the ViRL only model (ViRL Only) that is trained without any SFT data. 

\subsection{Comparison with task-specific models}\label{sec:task_specific}
We compare VinciCoder with other domain-specific models, like chart-to-code and image-to-SVG models.
The results in Table~\ref{tab:chartmimic_results} show that VinciCoder performs comparably to specialized models such as ChartCoder \cite{zhao2025chartcoder}, Chart2Code \cite{zhang2025boosting}, MSRL \cite{chen2025breaking} and ChartMaster \cite{tan2025chartmaster} on the ChartMimic dataset. More domain-specific models are compared in the Appendix. Moreover, its performance on UniSVG exceeds that of fine-tuned specialist models, which validates the effectiveness of our unified framework for multimodal code generation.

\begin{table}[t]
\setlength{\tabcolsep}{2pt}
\small
\centering
\resizebox{0.48\textwidth}{!}{
\begin{tabular}{l|ccc}
\toprule
\multirow{2}{*}{\makecell{\textbf{Chart-to-code} \\ \textbf{Models}}} & \multicolumn{3}{c}{\textbf{ChartMimic-direct-v2}} \\
\cmidrule{2-4}
& Exec. Rate & Low-Level  & High-Level \\
\midrule
ChartCoder & 91.4 & 72.5 & 74.0 \\
Chart2Code & 62.1 & 42.9 & 33.3 \\
MSRL & 96.5 & 78.6 & 83.8 \\
ChartMaster & 93.8 & 78.2 & 85.1 \\
\midrule
VinciCoder-7B  & 91.2 & 78.3 & 79.8 \\
VinciCoder-8B & 91.6 & 78.9 &  80.6 \\
\bottomrule
\end{tabular}}
\vspace{-5pt}
\caption{Comparing with task-specific chart-to-code models.}
\label{tab:chartmimic_results}
\vspace{-5pt}
\end{table}

\begin{table}[t]
\setlength{\tabcolsep}{2pt}
\small
\label{tab:task_results}
\centering
\resizebox{0.48\textwidth}{!}{
\begin{tabular}{l|cccc}
\toprule
\multirow{2}{*}{\makecell{\textbf{Image-to-SVG} \\ \textbf{Models}}} & \multicolumn{4}{c}{\textbf{UniSVG-ISVGEN}} \\
\cmidrule{2-5}
& SSIM $\uparrow$ & LPIPS $\downarrow$ & CLIP Score$\uparrow$ & Score \\
\midrule
LLaVA 1.5 Tuned & 65.4 & 47.9 & 80.2 & 71.6 \\ 
Llama 3.2 Tuned & 72.2 & 37.8 & 84.3 & 77.5 \\
Qwen2.5-VL Tuned & 72.5 & 36.8 & 83.6 & 77.3 \\

\midrule
VinciCoder-7B  & 77.4 & 23.3 & 92.0 & 86.0 \\
VinciCoder-8B & 77.4 & 23.2 & 94.1 & 87.3\\
\bottomrule
\end{tabular}}
\vspace{-5pt}
\caption{Comparing with task-specific image-to-SVG models. The metrics of finetuned models are from UniSVG \cite{li2025unisvg}.}
\vspace{-5pt}
\end{table}

\subsection{Effectiveness of Refinement Data}
We further evaluate the impact of refinement data by comparing models trained with and without its inclusion during the SFT stage. As shown in Table~\ref{tab:refinement_results}, incorporating these tasks yields consistent improvements across nearly all benchmarks. This observation suggests that exposure to refinement data during SFT not only enables iterative correction but also enhances the model's direct generation performance. These results underscore the effectiveness of refinement data in bridging the gap between initial code generation and final visual alignment.

\begin{table}[t]
\setlength{\tabcolsep}{2pt}
\small
\label{tab:refinment_results}
\centering
\resizebox{0.48\textwidth}{!}{
\begin{tabular}{l|cc|cc|cc}
\toprule
\multirow{2}{*}{Model} & \multicolumn{2}{c|}{\textbf{ChartMimic}} & \multicolumn{2}{c|}{\textbf{UniSVG}} & \multicolumn{2}{c}{\textbf{Design2Code}} \\
\cmidrule{2-7}
    & Low-L  & High-L & Low-L  & High-L   & Low-L & High-L \\ 
\midrule
w/o  Refine  & 76.2 & 76.4 & 75.6 & 89.4 & 86.1 & 86.6 \\
w/  Refine  & 75.8 & 78.6 & 78.2 & 91.0 & 86.5 & 87.0 \\
\bottomrule 
\end{tabular}
}
\vspace{-5pt}
\caption{Ablation studies about the 300k refinement data. We compare models after the SFT stage.}

\vspace{-10pt}
\end{table}

\subsection{ViRL Results}
\begin{figure}[t]
    \centering
    \includegraphics[width=0.45\textwidth]{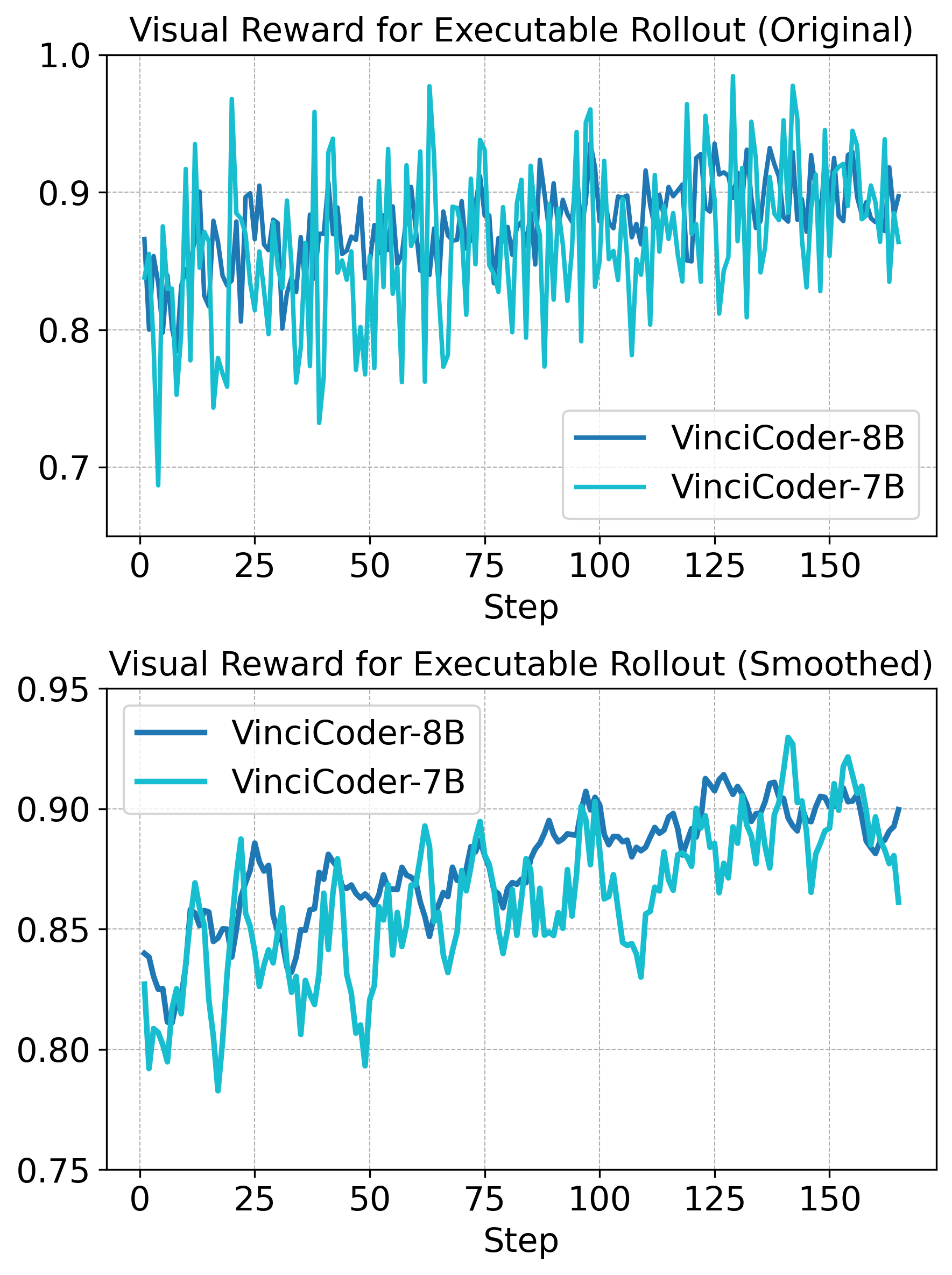}
    \vspace{-10pt}
    \caption{The reward progression of executable rollouts during our ViRL training stage.}
    \label{fig:rl_vis_exec}
    \vspace{-10pt}
\end{figure}

Figure~\ref{fig:rl_result} shows the improvement in visual similarity during RL training. Our reward function assigns a score based on visual similarity only for executable rollouts, while all unexecutable rollouts receive a reward of zero. We want to analyze whether the visual similarity between executable rollouts and input images improves during RL training. We visualize the visual reward of executable rollouts only during the RL training process. The result in Figure~\ref{fig:rl_vis_exec} denotes that during our training, the visual similarity of executable rollouts improves as well, demonstrating that our proposed ViRL training strategy not only improves the execution rate but also the visual fidelity.

\subsection{Fail Cases in ViRL}\label{sec:fail_case}
\begin{figure}[tbp]
    \centering
    \includegraphics[width=0.45\textwidth]{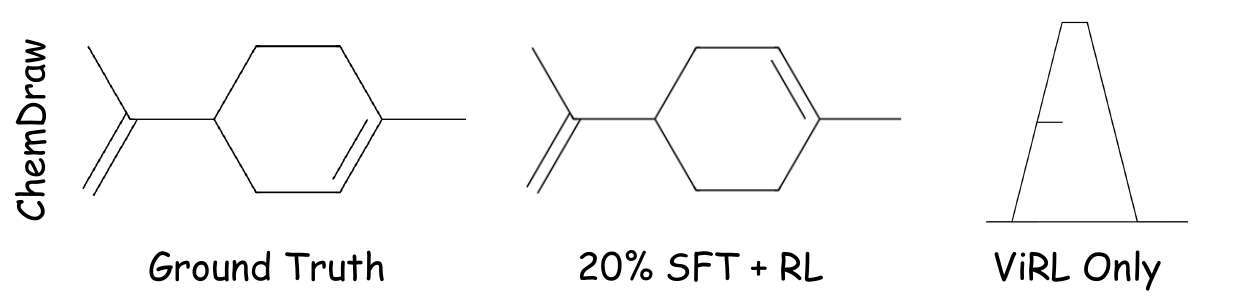}
    \vspace{-5pt}
    \caption{Example of semantically wrong code.}
    \label{fig:fail_chem}
    \vspace{-10pt}
\end{figure}

While ViRL enhances visual fidelity, it faces challenges with unfamiliar code structures. Specifically, the model can generate disorganized code that deviates from the input image and instructions, a behavior particularly evident in the ChemDraw benchmark. As illustrated in Figure~\ref{fig:fail_chem}, the ViRL Only model produces a geometric shape (a simple trapezoid) that bears no semantic relation to the chemical molecular structure in the ground truth. This suggests that without the prior knowledge instilled by SFT, RL alone may oversimplify the complex layout into primitive shapes that are easier to optimize but semantically incorrect. In contrast, the model trained with 20\% SFT + RL successfully reconstructs the intricate chemical bonds and rings, underscoring that even a small scale of SFT is essential for enforcing correct syntax and domain-specific code usage.

\section{Related Works about Multimodal Code Generation Benchmarks}
Besides constructing code MLLMs, many multimodal code generation benchmarks have been proposed for evaluation. Previous works generally focus on evaluating direct generation capacity within task-specific domains. For instance, benchmarks such as ChartMimic \cite{yang2024chartmimic} and Plot2Code \cite{wu2024plot2code} evaluate chart-to-code generation capabilities. Other works, including Design2Code \cite{si2024design2code}, UniSVG \cite{li2025unisvg}, and Image2Struct \cite{roberts2024image2struct}, assess the code generation of corresponding visual inputs. Also, some benchmarks evaluate beyond direct generation, including interaction \cite{xiao2024interaction2code, li2024sketch2code} and editing generation \cite{chen2025svgenius, zhao2025chartedit, yang2025chartm}.
Recently, with the growing capacities of VLMs, many new and complex benchmarks have been introduced. Artifactsbench \cite{zhang2025artifactsbench} and DCG-Bench \cite{li2025opusanimation} propose diverse webpage and chart code generation tasks with dynamic visual images and code complexity.

\section{GRPO Alogrithm}
A primary advantage of GRPO is its independence from a separate critic model, a key component in methods like PPO \cite{schulman2017proximal}. For a given input query $x$, the algorithm first samples a set of $G$ responses, $\{o_1, o_2, \dots, o_G\}$, from the current policy $\pi_{old}$. 
Each response receives a reward $R_i$ and the group-normalized advantage for the $i$-th response at time step $t$ is:
\begin{equation}
\small
\hat{A}_{i,t} = \frac{R_i - \text{mean}(\{R_j\}_{j=1}^G)}{\text{std}(\{R_j\}_{j=1}^G)}.
\end{equation}

It then optimizes the new policy $\pi_{\theta}$ by maximizing the following objective function:

\begin{equation}
\small
\label{eq:grpo}
\begin{aligned}
\mathcal{J}_\text{GRPO}(\theta) &=  \mathbb{E}_{(x,y)\sim\mathcal{D},\substack{\{o_i\}_{i=1}^G \sim \pi_{\theta_{\text{old}}} (\cdot \mid x)}} \\
&\biggl[ \frac{1}{G} \sum_{i=1}^G \frac{1}{|o_i|} \sum_{t=1}^{|o_i|} \biggl(\min \Bigl(r_{i,t}(\theta) \hat A_{i,t} \\
& \operatorname{clip}\left(r_{i,t}(\theta), 1-\varepsilon, 1+\varepsilon\right) \hat A_{i,t}\Bigl)  \biggr]
\end{aligned}
\end{equation}

\noindent where the probability ratio $r_{i,t}(\theta)$ is defined as:

\begin{equation}
\small
r_{i,t}(\theta) = \frac{\pi_\theta(o_{i,t} \mid x, o_{i,<t})}{\pi_{\theta_{\text{old}}}(o_{i,t} \mid x, o_{i,<t})}.
\end{equation}

\section{Refinement Task}
\begin{figure*}[h]
\begin{tcolorbox}[colback=white, colframe=black, title=Data Format for Refinement Task]

\textbf{Image (Ground Truth):}\\
\includegraphics[width=0.45\textwidth]{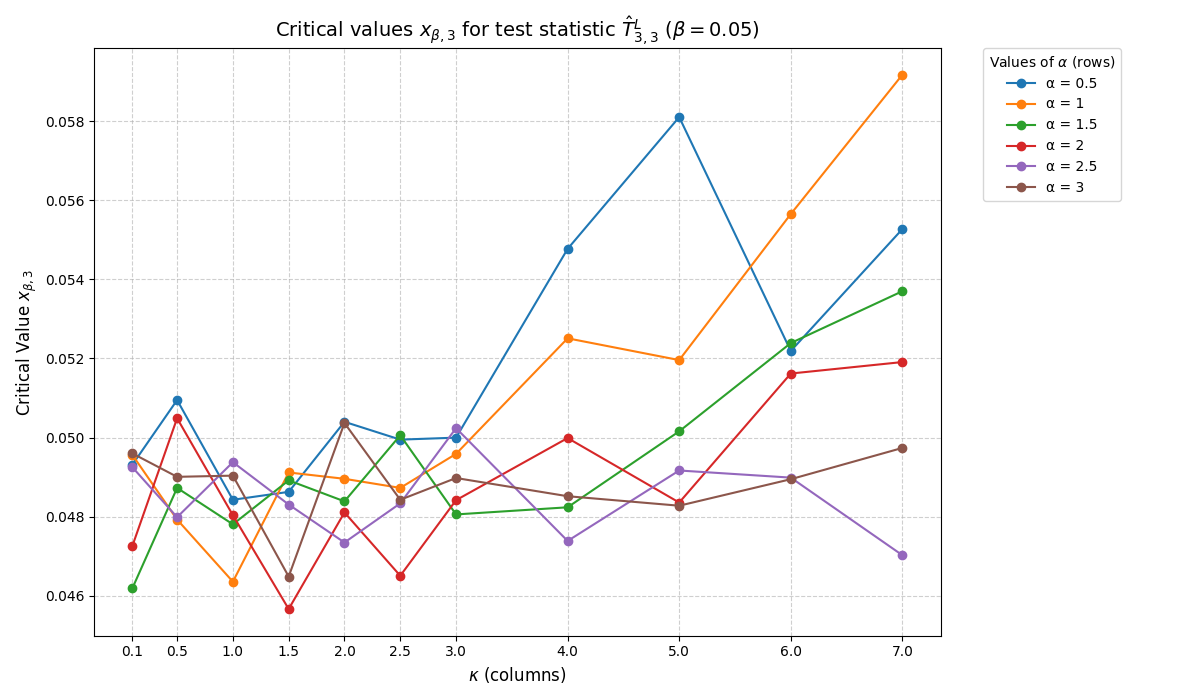}
\\
\textbf{User Instruction:}
You aim to generate a redrawn code of the given chart.
Previously, you generated an incomplete code implementation. Let's refine this plotting code. It seems to have inaccuracies and isn't working as expected. \\
Previous generated code: \\
{\color{blue} Incomplete Code Implementation} \\
Please modify the previous code snippets to ensure the code can render the given image. \\

\textbf{Assisstant Instruction:}\\
{\color{blue} Concrete Code Implementation}

\end{tcolorbox}
\caption{Data showcase of the visual-based refinement task.}
\label{fig:refine_task}
\end{figure*}

We provide the data format of the refinement task in Figure~\ref{fig:refine_task}. The incomplete code implementation is generated by our trained, dedicated VLM, and the concrete code implementation is from our curated datasets.

\section{Evaluation Prompts}

\begin{figure*}[t]
\begin{tcolorbox}[colback=white, colframe=black, title=ChemDraw Benchmark Prompts]

\textbf{Instruction:} Create a Python script that displays a visual representation of the molecule shown in the image. Utilize 'smiles = $<$SMILES$>$ string' to define the smile and RKiT or Indigo for plot.

\textbf{Image:}
\includegraphics[width=0.45\textwidth]{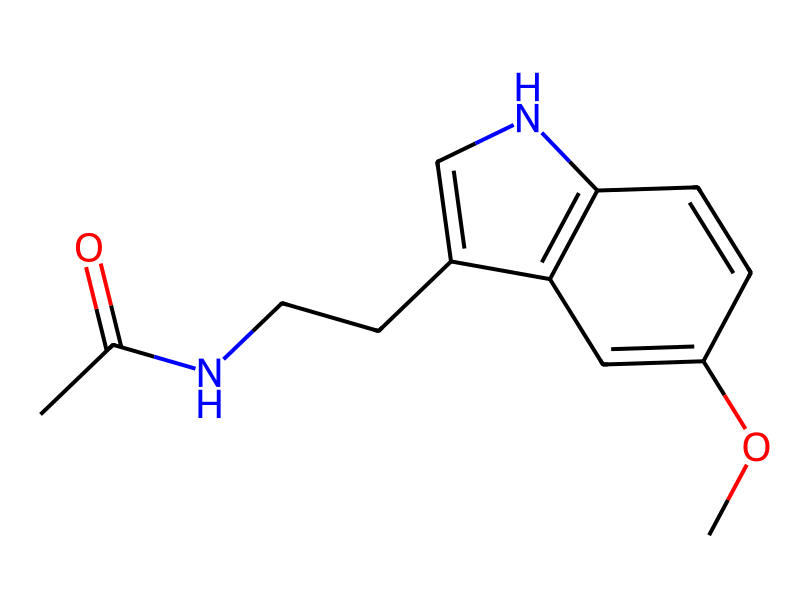}
\end{tcolorbox}
\caption{Prompt for our constructed ChemDraw benchmark.}
\label{fig:chemdraw}
\end{figure*}

In our experiments, all the prompts utilize the official implementations, except for our constructed  ChemDraw. For open-source models, we instruct them to generate output SMILES in the format of \texttt{'smiles = SMILES string'} and utilize robust parsers to extract the output SMILES for comparison. We visualize the prompt and the input molecule image in Figure~\ref{fig:chemdraw}.

\end{document}